%% file: main.tex
\documentclass{article}

\usepackage[utf8] {inputenc} %
\usepackage[T1]{fontenc}    %
\usepackage{hyperref}       %
\usepackage{url}            %
\usepackage{booktabs}       %
\usepackage{amsfonts}       %
\usepackage{nicefrac}       %
\usepackage{microtype}      %
\usepackage{xcolor}         %
\usepackage{xspace}
\usepackage{wrapfig}
\usepackage{multirow}
\usepackage{array}
\usepackage{caption,subcaption}
\usepackage{wrapfig}
\usepackage{lipsum, caption, booktabs, threeparttable}
\input{insbox.tex}

\usepackage{amsmath}
\usepackage{amssymb}
\usepackage{mathtools}
\usepackage{amsthm}
\usepackage{comment}
\usepackage{epsfig}
\usepackage{enumitem} 
\usepackage[bb=dsserif]{mathalpha} 
\usepackage{ragged2e}
\usepackage{balance}

\usepackage{color-edits} 
\addauthor{as}{red}
\addauthor{sm}{purple} 
\addauthor{zw}{purple}   

\usepackage[capitalize,noabbrev]{cleveref}

\usepackage[accepted]{icml2024}

\usepackage{amsmath}
\usepackage{amssymb}
\usepackage{mathtools}
\usepackage{amsthm}
\usepackage{comment}
\usepackage{xcolor}
\usepackage{xspace}

\usepackage[capitalize,noabbrev]{cleveref}

\theoremstyle{plain}

\theoremstyle{definition}

\theoremstyle{remark}

\newcommand{\R}{\mathbb{R}}

\usepackage[textsize=tiny]{todonotes}

\input{project_macros}

\icmltitlerunning{Random Latent Exploration for Deep Reinforcement Learning} 

\begin{document}

\twocolumn[
\icmltitle{Random Latent Exploration for Deep Reinforcement Learning
}

\icmlsetsymbol{equal}{*}

\begin{icmlauthorlist}
\icmlauthor{Srinath Mahankali}{imp,yyy}
\icmlauthor{Zhang-Wei Hong}{imp,yyy}
\icmlauthor{Ayush Sekhari}{yyy} 
\icmlauthor{Alexander Rakhlin}{yyy}
\icmlauthor{Pulkit Agrawal}{imp,yyy} 
\end{icmlauthorlist}

\icmlaffiliation{imp}{Improbable AI Lab}
\icmlaffiliation{yyy}{Massachusetts Institute of Technology}

\icmlcorrespondingauthor{Srinath Mahankali}{srinathm@mit.edu} 

\icmlkeywords{Machine Learning, ICML}

\vskip 0.3in
]

\renewcommand{\thefootnote}{\relax}
\printAffiliationsAndNotice{} %

\begin{abstract}
We introduce Random Latent Exploration (RLE), a simple yet effective exploration strategy in reinforcement learning (RL). On average, RLE outperforms noise-based methods, which perturb the agent's actions, and bonus-based exploration, which rewards the agent for attempting novel behaviors. The core idea of RLE is to encourage the agent to explore different parts of the environment by pursuing randomly sampled goals in a latent space. RLE is as simple as noise-based methods, as it avoids complex bonus calculations but retains the deep exploration benefits of bonus-based methods. Our experiments show that RLE improves performance on average in both discrete (e.g., Atari) and continuous control tasks (e.g., Isaac Gym), enhancing exploration while remaining a simple and general plug-in for existing RL algorithms. Project website and code at \url{https://srinathm1359.github.io/random-latent-exploration}.

\end{abstract} 

\section{Introduction}
\label{sec:intro}
\input{new_intro.tex}

\section{Preliminaries}
\label{sec:prelim}
\paragraph{Reinforcement Learning (RL).} RL is a popular paradigm for solving sequential decision-making problems  \citep{sutton2018reinforcement} where an agent operates in an unknown  environment~\cite {sutton2018reinforcement} and aims to improve its performance through repeated interactions with the environment. At each round of interaction,  the agent starts from an initial state $s_0$ of the environment and collects a trajectory. At each timestep $t$ within that trajectory, the agent perceives the state $s_t$, takes an action $a_t \sim \pi(.|s_t)$ with its policy $\pi$, receives a task reward $r_t = r(s_t, a_t)$, and transitions to a next state $s_{t+1}$ until reaching terminal states,  after which a new trajectory is initialized from \(s_0\) and the above repeats. The goal of the agent is to learn a policy $\pi$ that maximizes expected return $\mathbb{E}_{\pi}\Big[\sum^{\infty}_{t=0} \gamma^{t} r(s_t, a_t) \Big]$. A straightforward approach is to estimate the expected return of a policy by rolling out trajectories $(s_0, a_0, s_1, \cdots, )$ through Monte Carlo sampling \citep{konda1999actor}, and then optimizing this to find the optimal policy, but unfortunately, the corresponding estimates are of high variance and thus often require a huge number of data. Thus, in practice, various RL algorithms learn a value function (or value network) $V^{\pi}$ from the interaction that approximates 
\begin{align}
    V^{\pi}(s_0) \approx \mathbb{E}_{\pi}\Big[\sum^{\infty}_{t=0} \gamma^{t} r(s_t, a_t) \Big],
\end{align}
and train the policy $\pi$ to maximize the value $V^\pi(s_0)$ (e.g. using policy gradient). 

\textbf{Exploration.} As the reward may be delayed and not immediately presented to the agent, the agent may need to take many actions and visit a sequence of states without rewards before it receives any learning signal (reward). As such, taking greedy action $a_t$ at each step that maximizes immediate reward $r(s_t,a_t)$ does not necessarily lead to a high return. Thus, RL algorithms require ``exploring'' states and actions that may lead to low immediate rewards but could potentially end up with high return in the long run.  We refer to this process as \textit{exploration} throughout this paper~\citep{sutton2018reinforcement}.

\section{Our Method: Random Latent Exploration} 

\textbf{Problem statement.} 
We aim to develop an easy-to-implement exploration strategy that improves over the standard action noise exploration in discrete and continuous action spaces.

\textbf{Challenge: Ensuring diversity.} 
The key challenge in noise-based exploration is their limited ability to generate diverse behaviors. Although noise-based exploration changes the agent's actions at each timestep, in practice these perturbations are too local to deviate significantly from the usual trajectories. As a result, the generated trajectories can appear similar to each other, limiting exploration to a narrow region of the environment. The primary technical challenge, then, is devising a method that ensures such diversity in the generated trajectories.

\textbf{Approach: Randomized rewards.} 
To overcome the above challenge, we propose enhancing trajectory diversity by altering the agent's rewards, inspired by skill discovery methods \citep{eysenbach2018diversity} that learn skills based on different reward functions that push the agent to visit different parts of the state space for different skills. Following a similar idea, in every training episode, we perturb the given task reward function by adding a randomized state-dependent reward function, and we train policies to maximize the sum of randomized and task rewards. The key idea is that in every round of interaction, the agent is incentivized to visit areas of the state space with high randomized rewards, and if these random areas are diverse enough, we will get diverse behaviors in the environment during training---thus incentivizing exploration. However, since the random rewards are repeatedly resampled and thus keep on changing during training to ensure stable and effective learning, both policies and value functions must be aware of the specific random reward function in use; otherwise, the changing reward functions will comprise a partially observable MDP \citep{kaelbling1998planning}. We take inspiration from the Universal Value Function Approximator (\textsc{UVFA}) \citep{schaul2015universal}, which trains networks based on varying goals. We adopt their approach by equating their goals with different reward functions, thus making the policy $\pi$ and the value function $V^\pi$ condition on the sampled reward function. This ensures that the random rewards no longer appear to be noise to the policy. The remaining questions are: 
\begin{itemize}[leftmargin=*, itemsep=1mm] 
    \item How to implement the randomized reward functions?
    \item How to make the policy and the value function condition on the sampled reward functions? 
\end{itemize}
In the next section, we outline our implementation of the above idea, and the full implementation is provided in \Cref{app:impl}.

\subsection{Algorithmic Implementation}
\label{subsec:method:alg}
\textbf{Randomized reward functions.} 
We design a practical approach to implement randomized reward functions using two principles. First, the randomized reward function must depend on the observation; otherwise the rewards would be random noise, which does not help exploration, as shown by \citet{fortunato2017noisy}. Second, the policy and value functions must be conditioned on the randomized reward function so that the reward function is fully observable to them. Following these criteria, we efficiently implement the randomized reward function as the dot product of the state feature and a randomly chosen latent vector:
\begin{align}
\label{eq:method:Z}
    F(s, \boldsymbol{z}) = \phi(s) \cdot \boldsymbol{z},
\end{align}
where \(\phi: \mathcal{S} \to \mathbb{R}^d\) is a feature extractor that transforms a state into a \(d\)-dimensional vector, and \(\boldsymbol{z} \in \mathbb{R}^d\) represents a latent vector. Randomized rewards for each state are generated by sampling \(\boldsymbol{z}\) from a given distribution \(P_{\boldsymbol{z}}\), and then setting rewards as \(F(s, \boldsymbol{z})\). Even if $F(s,\boldsymbol{z})$ is high at unreachable parts of the state space, the agent simply does not collect the reward and moves on. This does not derail training as $\boldsymbol{z}$ is resampled at the start of each trajectory.

\textbf{Latent conditioned policy and value network.} 
Recall that the policy and the value function must be aware of the state and the random variable that factorizes the randomized reward function $F$. To achieve this, we augment the input to the policy and the value functions with the latent vector $\boldsymbol{z}$. The resulting policy and the value networks are $\pi(.|s, \boldsymbol{z})$ and $V^{\pi}(s, \boldsymbol{z})$. We train the latent-conditioned value network to approximate the expected sum of the original reward and the randomized rewards as below 
\begin{align}
    V^{\pi}(s, \boldsymbol{z}) \approx \mathbb{E}_{\pi}\left[ \sum^{\infty}_{t=0}\gamma^{t} (R(s_t, a_t) + \lambda F(s_{t+1}, \boldsymbol{z})) \right],
\end{align}
where $\lambda$ is a hyperparameter. We train the latent-conditioned policy $\pi$ to maximize $V^{\pi}(s, \boldsymbol{z})$ at every state $s$ and latent vector $\boldsymbol{z}$. Both value and policy networks can be trained with any off-the-shelf RL algorithms, e.g. PPO~\citep{schulman2017proximal}, DQN~\citep{mnih2015human}, A3C~\citep{mnih2016asynchronous}, SAC~\citep{haarnoja2018soft}.

\textbf{Latent vector sampling.} To ensure the agent is exposed to a wide range of randomized reward functions, we randomize the latent vector $\boldsymbol{z}$, which is done by resampling at the start of each trajectory. This ensures each trajectory is rolled out under the same policy and latent vector $\boldsymbol{z}$, maintaining temporal consistency crucial for deep exploration, as indicated by prior work \citep{osband2016deep,fortunato2017noisy}. The sampling distribution of $\boldsymbol{z}$ is discussed in Section~\ref{sec:exp}.

As we train the policy conditioned on the randomly sampled latent $\boldsymbol{z}$ that defines the randomized reward function, we term our method as \textit{Random Latent Exploration (RLE)}. We outline the algorithm in Algorithm \ref{alg:alg} and present the detailed version in \Cref{app:alg:alg} (in Appendix). Note that at line 6 in  Algorithm \ref{alg:alg}, we compute the randomized reward using the next state $s_{t+1}$ since the next state reflects the effect of the agent's chosen action $a_t$ in state $s_t$. This choice is common in prior works computing exploration bonuses, where the exploration bonus is a function of the newly-reached state rather than the current state~\citep{burda2018exploration}.

\begin{algorithm}[H]
\begin{algorithmic}[1]
    \STATE \textbf{Input:} Latent distribution $P_{\boldsymbol{z}}$ 
    \REPEAT 
        \STATE Sample a fresh latent vector: $\boldsymbol{z} \sim P_{\boldsymbol{z}}$ 
        \FOR {$t = 0, \dots, T$} 
            \STATE Take action $a_t \sim \pi(.| s_t, \boldsymbol{z})$ and transition to $s_{t+1}$ 
            \STATE  Receive reward: $r_t = R(s_t, a_t) + F(s_{t+1}, \boldsymbol{z})$
        \ENDFOR
        \STATE Update policy network $\pi$ and value network $V^\pi$ with the collected trajectory $(\boldsymbol{z}, s_0, a_0, r_0, s_1, \cdots, s_{T})$
    \UNTIL convergence       
\end{algorithmic}
    \caption{Random Latent Exploration (\RLE)} 
    \label{alg:alg} 
\end{algorithm}

\section{Experiments} 
\label{sec:exp}

We compare \RLE performance against the action noise exploration method typically used in many RL algorithms \citep{schulman2017proximal,mnih2015human}. In all our experiments, we train the agent using \PPO \citep{schulman2017proximal} for each task separately. Standard \PPO implementation \citep{schulman2017proximal} explores by sampling actions from the learned policy (i.e., a Boltzmann distribution over actions for discrete action spaces or a Gaussian distribution for continuous action spaces). 

We also compare \RLE  with the following  exploration strategies as baselines:
\begin{itemize}[leftmargin=*]
    \item \textbf{\NoisyNet} \citep{fortunato2017noisy}: We chose it to be the representative baseline from the family of noise-based exploration \citep{osband2016deep,fortunato2017noisy,plappert2017parameter} because it has been used in prior works on benchmarking exploration strategies and achieve superior performance\citep{chen2022redeeming,taiga2019benchmarking}.
    \item \textbf{\RND} \citep{burda2018exploration}: We choose \RND to be the representative baseline from the family of bonus-based exploration methods since it shows considerable improvements over action noises and noise-based approaches in hard-exploration tasks in \Atari. 
\end{itemize}

\subsection{Illustrative Experiments on \textsc{FourRoom}}
\label{subsec:exp:illustrative}
We first ran toy experiments on the \FourRoom environment~\citep{sutton1999between} to test whether our method can perform deep exploration.

\begin{wrapfigure}{r}{0.4\columnwidth}
  \vspace{-4ex}
  \begin{center}
    \includegraphics[width=0.4\columnwidth, trim={0.2cm 0.3cm 0.3cm 0.2cm},clip]{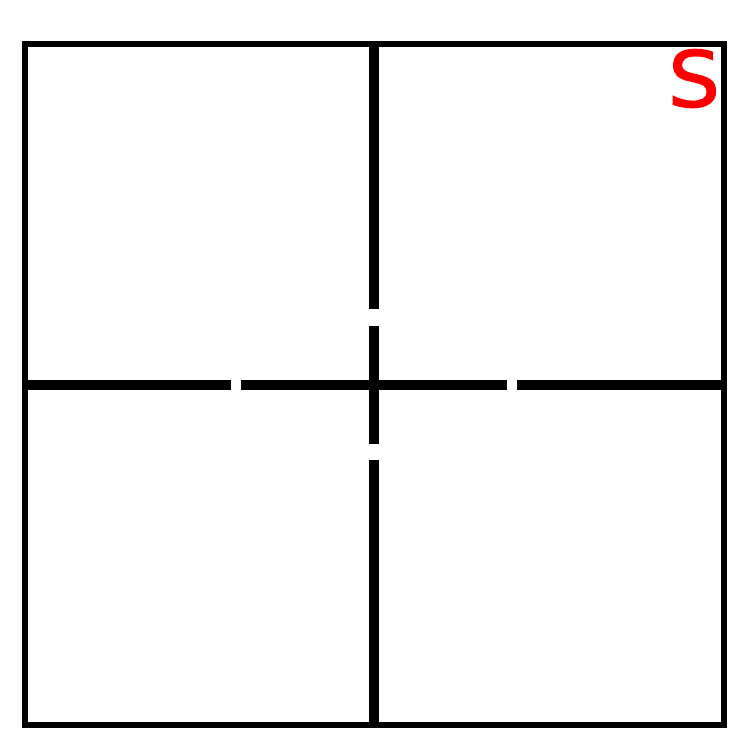}
  \end{center}
  \caption{\FourRoom environment. The agent starts at the top-right state (denoted by red `S') and can move left, right, up, and down. The black bars denote walls that block the agent's movement.}
  \label{fig:fourroom_setup}
  \vspace{-1.8ex} 
\end{wrapfigure}

\textbf{Setup.} Figure \ref{fig:fourroom_setup} illustrates \FourRoom environment with $50 \times 50$ states consisting of four rooms separated by solid walls (which the agent can't cross) and connected with small openings of a single state each (which the agent needs to go through to travel across rooms).
The agent observes the $(x,y)$ coordinates as the state and can take action to move left, right, up, and down (if not interrupted by a wall). At the beginning of each trajectory, the agent always starts from the top-right corner of the room (denoted by the letter ``S''). In this study, we always give zero task reward to the agent since we are interested in comparing the deep exploration behavior of different strategies. This is also known as reward-free exploration~\cite{pathak2017curiosity}.\footnote{We also perform experiments on \FourRoom with an sparse task reward of \(1\) at the bottom-left corner. The results and visualization of visitation counts are deferred to Appendix \ref{app:fourroom_visualizations}.}  

\setlength{\columnsep}{8pt}

We compared agents trained with different exploration strategies: \PPO, \NoisyNet, \RND, and \RLE, over 2.5 million timesteps. For \RLE, the feature extractor defined in \Cref{eq:method:Z} is set to be a randomly initialized neural network with one hidden layer, the output layer of which has the same dimension as $\boldsymbol{z} \sim P_{\boldsymbol{z}}$ to make the final dot product operation feasible. Further implementation details are available in \Cref{app:four_room_details}.  Because of the walls, the \FourRoom environment requires deep exploration to go to states distant from the initial states.

\begin{figure}[t!] 
    \centering
    \includegraphics[width=0.678\columnwidth, trim={0.2cm 0.3cm 0.3cm -0.23cm},clip]{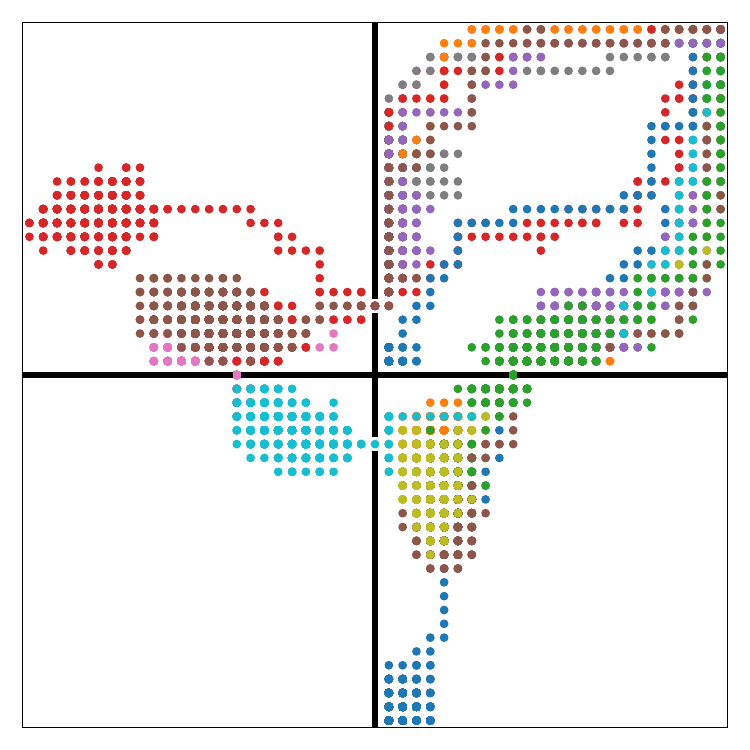}
    \caption{Rollout of multiple trajectories from a policy trained with \RLE in the middle of the training (1.5 million timesteps), where each color denotes a distinct trajectory. As the figure demonstrates, changing the latent vector $\boldsymbol{z}$ in \RLE leads to diverse trajectories across all four rooms.} 
    \label{fig:Trajectory_Diversity1}  
\end{figure}

\textbf{Does \RLE qualitatively improve trajectory diversity during training?}
The exploration of \RLE is driven by sampling the latent vector $\boldsymbol{z}$ to change the behavior of the policy network $\pi$ and the objective of the value network $V^\pi$. To investigate this, one may wonder how the choice of $\boldsymbol{z}$ affects the behaviors of the induced policy. To qualitatively understand this aspect, in Figure~\ref{fig:Trajectory_Diversity1} we plotted different trajectories corresponding to different choices of the latent vector $\boldsymbol{z} \sim P_z$. Each trajectory corresponds to a specific choice of $\boldsymbol{z}$ and is assigned a unique color. For this plot, we chose the checkpoint of the policy network stored in the middle of training (i.e., 1.5 million timesteps) to observe the trajectory diversity of the agent as it explores during training. Figure~\ref{fig:Trajectory_Diversity1} shows that the agent's exploration is diverse and visits all four rooms. This simple experiment furthers our belief that diverse choices of latent vector $\boldsymbol{z}$ can induce diverse trajectories. Similar plots over more seeds for all the exploration algorithms \RLE, \PPO, \RND, and \NoisyNet are provided in Appendix \ref{app:fourroom_visualizations}. 

\textbf{Explaining the observed trajectory diversity.}
To explain why the trajectories generated by an \RLE policy are diverse, we plot reward functions induced by sampling different $\boldsymbol{z}$ in \Cref{app:fig:random_reward1} (in Appendix). These plots demonstrate the diversity of the random rewards, each of which can guide the policy to a different part of the state space.

\begin{figure}[t!]
    \centering
    \includegraphics[width=0.72\columnwidth,clip]{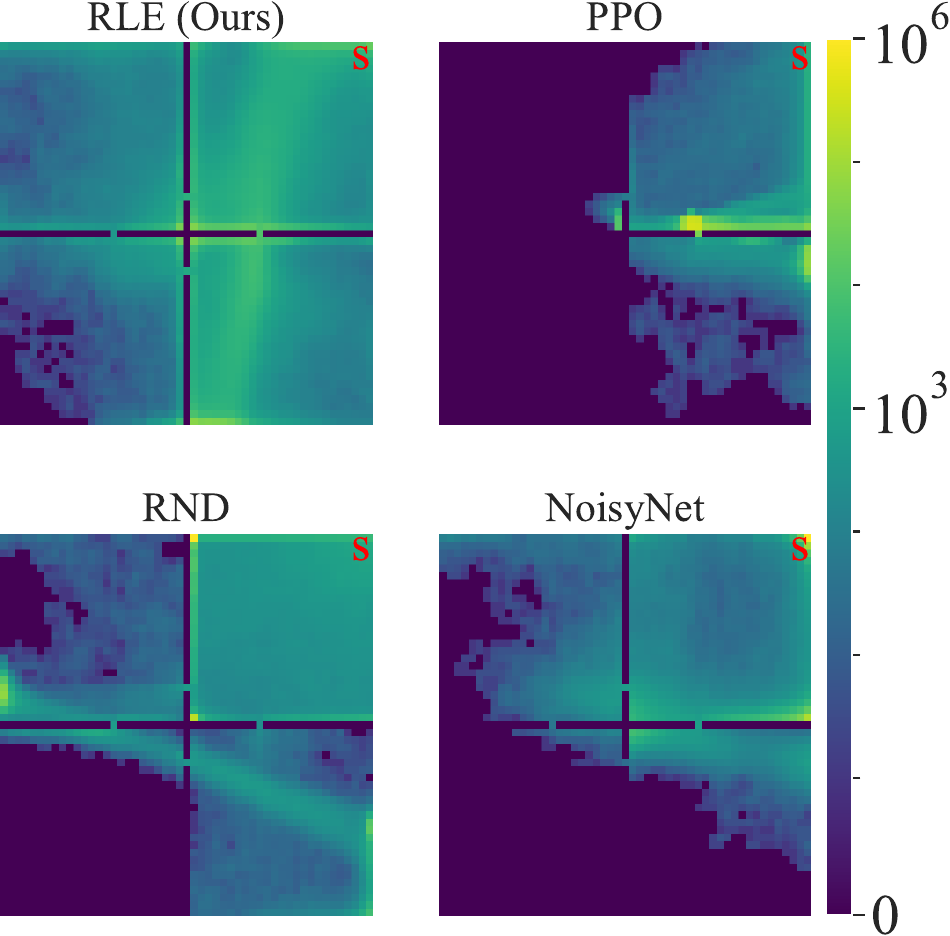}
    \caption{ State visitation counts of all the methods after training for 2.5M timesteps without any task reward (reward-free exploration). The start location is represented by the  red `S' at the top right. \RLE achieves much wider state visitation coverage over the course of training compared to other baselines, confirming that the diverse trajectories generated by the policy are useful for exploration. 
    } 
    \label{fig:visitation1} 
\end{figure}

\textbf{Quantifying State Visitation}
The state visitation counts of different exploration strategies is visualized in Figure~\ref{fig:visitation1}. The results show that PPO's state visitation centers around the starting room (i.e., top-right), indicating that action noise alone doesn't encourage the agent to explore far from the initial state. In contrast, \RLE, \RND, and \NoisyNet can frequently reach the rooms beyond the initial room, with \RLE visitation count spread across the four rooms. This suggests that RLE can do deep exploration similar to prior deep exploration algorithms for this environment.

\subsection{Benchmarking Results on \textsc{Atari}}
\label{subsec:exp:result}
Having performed illustrative experiments on the \FourRoom toy environment, we now evaluate our method on more realistic and challenging tasks. Our results below show that \RLE based exploration improves PPO's overall performance on most tasks. 

\textbf{Setup.} We evaluate our method in the well-known \Atari benchmark \citep{bellemare2013arcade}. Following the common practice in \Atari \citep{mnih2015human}, the agent observes a stack of the most recent four $84\times 84$ grayscale frames as input and outputs one of many discrete actions available in the \Atari (see \citet{bellemare2013arcade} for further environment details). For \RLE, we chose the feature learned by the value network followed by a randomly initialized linear layer as $\phi$ (used in Equation~\ref{eq:method:Z}). Note that the randomly initialized linear layer is kept frozen throughout training. We set the dimension of the latent vector \(\boldsymbol{z}\) as $8$. We test different values of the dimension of $\boldsymbol{z}$ in \Cref{subsec:exp:ablation} on a subset of games and observe similar performance. We also test using a completely randomly initialized network as $\phi$ in \Cref{app:sec:atari_detailed} and observe slightly worse performance.

We use the standard PPO hyperparameters~\citep{burda2018exploration} for training, and all implementation details are provided in Appendix \ref{app:impl}. For each \Atari game (i.e., environment), we train five policies with five different random seeds for $40$ million frames each following prior work \citep{chen2022redeeming,bellemare2016unifying}. However, we trained \textsc{Montezuma's Revenge} 
for $200$ million frames since its exploration difficulty is much harder than other \Atari games \citep{burda2018exploration}.

\begin{figure}[t!]
    \centering
    \includegraphics[width=0.4\textwidth]{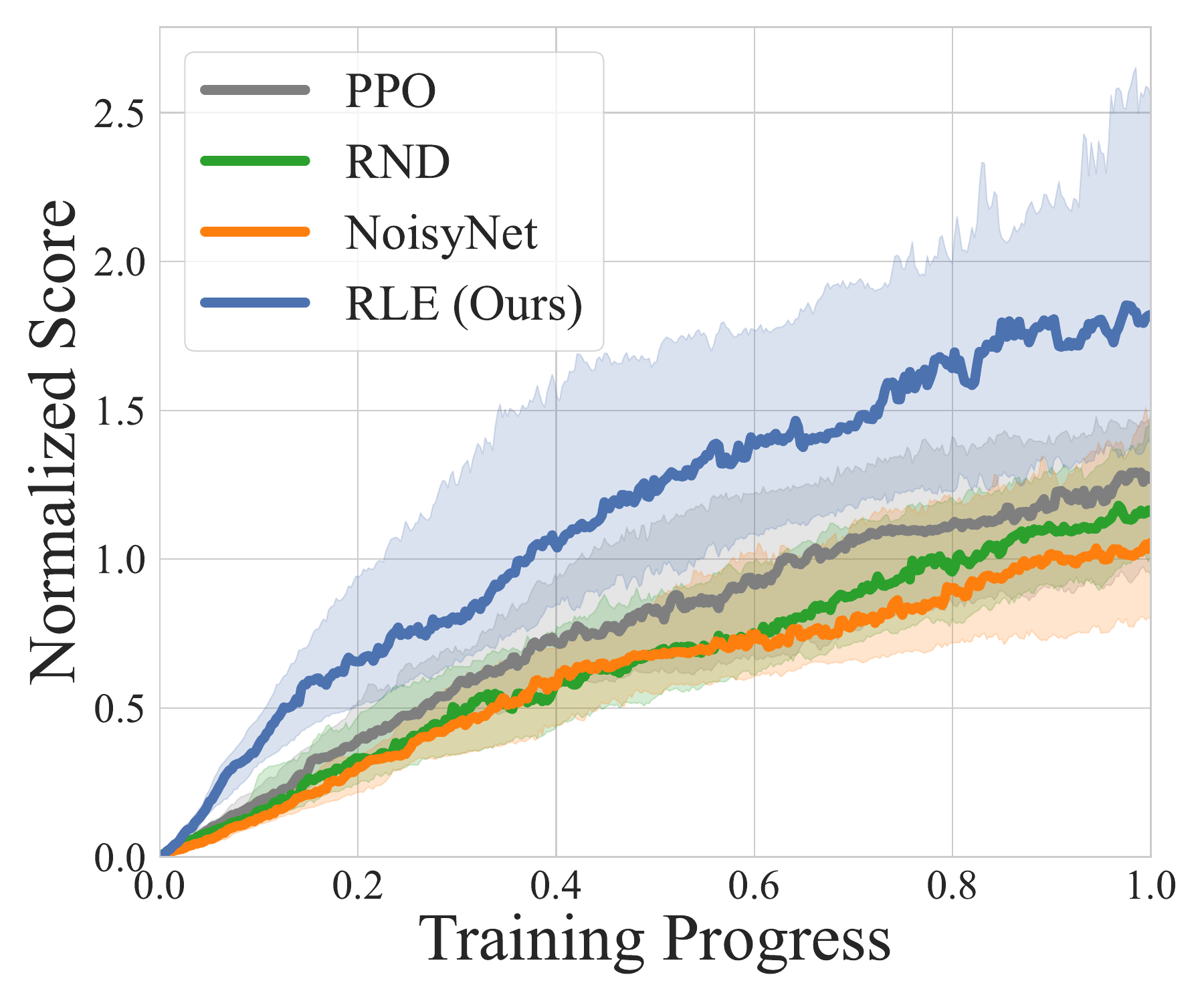}
    \caption{
Aggregated human normalized score across all 57 ATARI games. \RLE exhibits a higher interquartile mean (IQM) of normalized score than PPO across 57 ATARI games, showing that \RLE improves over PPO in the majority of tasks.}
    \label{fig:result:all}
\end{figure}

\textbf{Does \RLE improve the overall performance?} We answer this question by calculating the interquartile mean (IQM) \citep{agarwal2021deep} and its 95\% confidence interval, which was estimated using the bootstrapping method \citep{diciccio1996bootstrap} on the aggregated human-normalized scores from 57 games. Unlike empirical mean scores, IQM mitigates the influence of outliers on the aggregated metric. Figure~\ref{fig:result:all} demonstrates that \RLE achieves a higher IQM human-normalized score compared to all baselines, indicating that \RLE enhances performance over other exploration strategies in the majority of \Atari tasks. Besides the aggregate results, we present the learning curves for all methods across the 57 \Atari games in \Cref{fig:atari_learning_curves}. Additionally, the final mean score of each method across five seeds for each \Atari game is provided in \Cref{tab:atari_results} in the Appendix. From \Cref{fig:atari_learning_curves}, we notice that \RLE does not perform well on \textsc{Montezuma's Revenge}, indicating that while \RLE assists in producing diverse trajectories, \textsc{Montezuma's Revenge} still presents a challenge when not relying on bonus-based exploration.

\textbf{Does \RLE improve over the baselines consistently?} The aggregated performance improvement across all games reported in the previous Section doesn't reveal how many games \RLE outperforms the baselines. If \RLE outperforms baselines on almost all games, where each game can be thought as a different MDP, then we could say that with high-probability \RLE will win against other algorithms on new MDPs/tasks. To evaluate this, 
we measure the probability of improvement (POI;~\citet{agarwal2021deep}) between algorithms and their 95\% confidence intervals, estimated using the bootstrapping method \citep{diciccio1996bootstrap} which is reported in~\Cref{fig:result:poi_over_baselines} and~\Cref{fig:result:poi_over_ppo} (in Appendix).  
\Cref{fig:result:poi_over_baselines}(a) shows that the lower confidence bound of POI for \RLE over each algorithm is above 0.5, indicating that \RLE statistically outperforms the other baselines \citep{agarwal2021deep}. 
This means that for a randomly chosen task in \Atari, running \RLE is likely to yield a higher score than the other baselines, implying that \RLE's performance improvements are consistent and not limited to a few games. 

Conversely, \Cref{fig:result:poi_over_ppo}(a) reveals that the POI over \PPO for both \NoisyNet and \RND is below 0.5, suggesting that \NoisyNet and \RND do not consistently improve over \PPO despite having better performance in a few games (see Figure~\ref{fig:atari_learning_curves}). 
We use a CNN to represent the policy and value function in all \Atari experiments, rather than an LSTM as done in~\citep{chen2022redeeming} as the CNN-based architectures were also used in prior work~\citep{burda2018exploration} and due to its simplicity in implementation. This could explain why \PPO outperforms \RND on average over the 57 \Atari games in our experiments.

\begin{figure*}[ht!]
    \centering
    \includegraphics[width=0.9\textwidth]{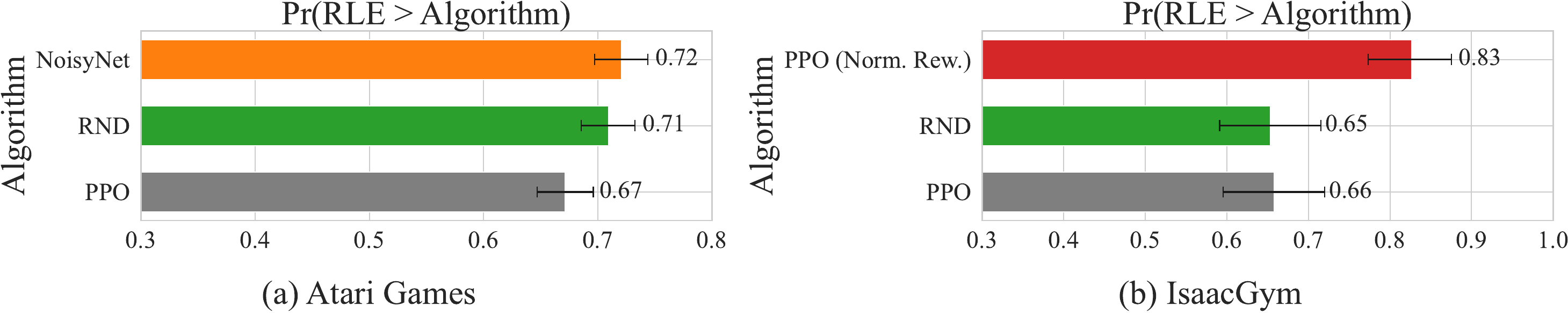}
    \caption{
    \textbf{(a)} Probability of improvement (POI) of our method, \RLE, over the baselines \NoisyNet, \RND and \PPO across all $57$ \Atari games (higher is better). The lower confidence bound of \RLE's POI over the other algorithms are all greater than $0.5$. This means that \RLE statistically improves over other algorithms \citep{agarwal2021deep}.
    \textbf{(b)} Probability of improvement of \RLE over the baselines \RND, \PPO, and \PPO with reward normalization across all $9$ \IsaacGym tasks. In this domain as well, the  lower confidence bound of \RLE's POI over the other algorithms are all greater than $0.5$. This means \RLE statistically improves over the other algorithms.
    }
    \label{fig:result:poi_over_baselines}
\end{figure*}

\begin{figure*}[tb!]
    \centering
    \includegraphics[width=0.95\textwidth]{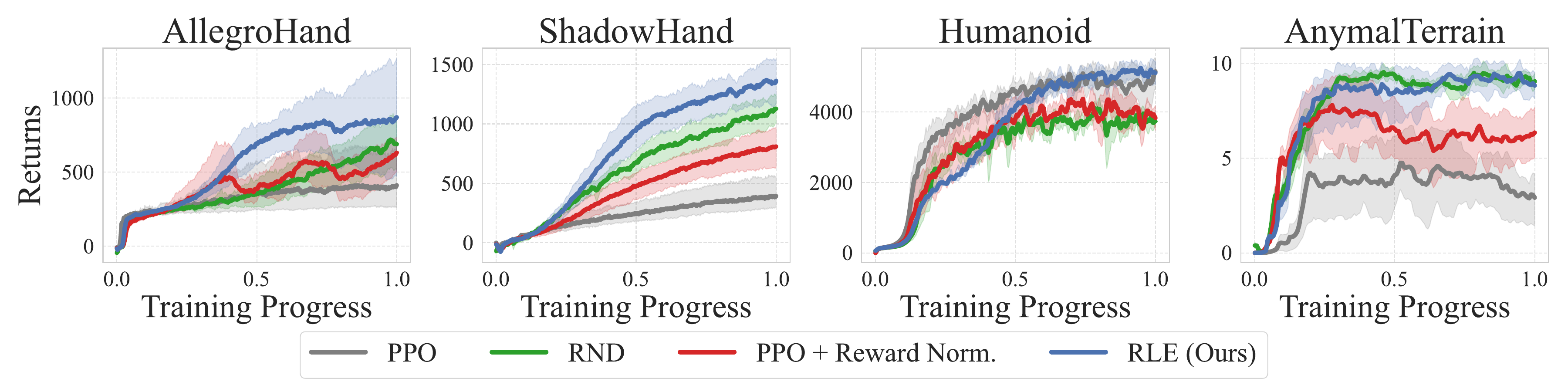}
    \caption{Comparison of performance between \RLE, \PPO, and \RND in four \IsaacGym tasks (higher is better). \RLE achieves higher return than \PPO and \RND in the majority of tasks, especially in tasks like \textsc{AllegroHand} and \textsc{ShadowHand} that require more exploration. This suggests \RLE improves over \PPO and \RND in continuous control domains as well.} 
    \label{fig:result:four_isaacgym}
\end{figure*}

\subsection{Evaluation on \textbf{\textsc{Isaac Gym}}} 
\label{subsec:exp:result_isaac}
To demonstrate that \RLE can improve upon \PPO in both discrete and continuous control tasks, we also conducted experiments in \IsaacGym \citep{makoviychuk2021isaac}, a benchmark suite containing numerous continuous control tasks. We compare the performance of \RLE against \PPO and \RND. As \RLE and \RND both use a secondary reward function with potentially different scale from the task reward function, we use reward normalization for these methods~\cite{burda2018exploration}. Thus, we compare against \PPO with reward normalization as an additional baseline.

We implemented \RLE on top of \PPO and trained it with standard \PPO hyperparameters in \IsaacGym provided in \texttt{CleanRL}, with implementation details provided in \Cref{subsec:isaacgym_details}. The performance of four representative tasks and all the tasks is reported in \Cref{fig:result:four_isaacgym} and~\Cref{fig:result:all_isaacgym} (in Appendix) respectively.

\textbf{Does \RLE improve over baselines in continuous control?} The results show that \RLE achieves a higher average return than \PPO in most tasks, with particularly large performance gains in \textsc{AllegroHand} and \textsc{ShadowHand}, indicating that \RLE improves upon \PPO in continuous control tasks. In \textsc{AllegroHand} and \textsc{ShadowHand}, the objective is to control an anthropomorphic hand to reorient objects to a target pose. These tasks require more exploration than other continuous control tasks since it takes many steps to achieve the target pose. To study the overall performance, we measure the probability of improvement of \RLE over \PPO and \RND across nine different \IsaacGym tasks and also present the results in \Cref{fig:result:poi_over_baselines} and \Cref{fig:result:poi_over_ppo}. \Cref{fig:result:poi_over_baselines}(b) shows that \RLE has a statistically significant chance of improving over both \PPO and \RND in \IsaacGym tasks as the lower confidence bound of the POI for \RLE over each baseline method is greater than $0.5$. Furthermore, \Cref{fig:result:poi_over_ppo}(b) shows that out of all considered methods, \RLE is the only one with a statistically significant POI over \PPO. To estimate the average performance difference across environments, we also measure the aggregated normalized return of \RLE, \PPO, and \RND across all \IsaacGym tasks. As the return is positive for \PPO in each environment, we normalize runs by dividing by the mean score of \PPO in that environment (i.e., normalize \PPO to have a score of 1). For further details of this metric, see \Cref{app:isaac_eval_details}. We present the results in \Cref{fig:result:iqm_normalized_score_isaacgym} and \Cref{fig:result:mean_normalized_score_isaacgym}, which show that \RLE achieves higher aggregate performance than \PPO and matches \RND in this metric.

\subsection{Ablation Studies}
\label{subsec:exp:ablation}
We ablated various design choices in \RLE using both the \(\Atari\) and \(\IsaacGym\) benchmarks.

\textbf{Effect of latent vector distribution.}
We investigated the impact of different latent vector distributions on \RLE's performance by choosing different sampling strategies including  $\mathrm{Uniform}([-0.5, 0.5]^d)$ and isotropic normal $\mathcal{N}(\boldsymbol{0}, \mathbf{I}_d)$ distributions, within a $d$-dimensional space where \(d = 8\). The detailed implementation is described in Appendix~\ref{app:impl}. The results presented in Figure~\ref{fig:ablation:dist} indicate that \RLE performs better than \PPO across different latent vector distributions. This suggests that \RLE's efficacy is not significantly affected by the choice of latent vector distribution.

\begin{figure}[htb!] 
    \centering
    \includegraphics[width=0.5\textwidth]{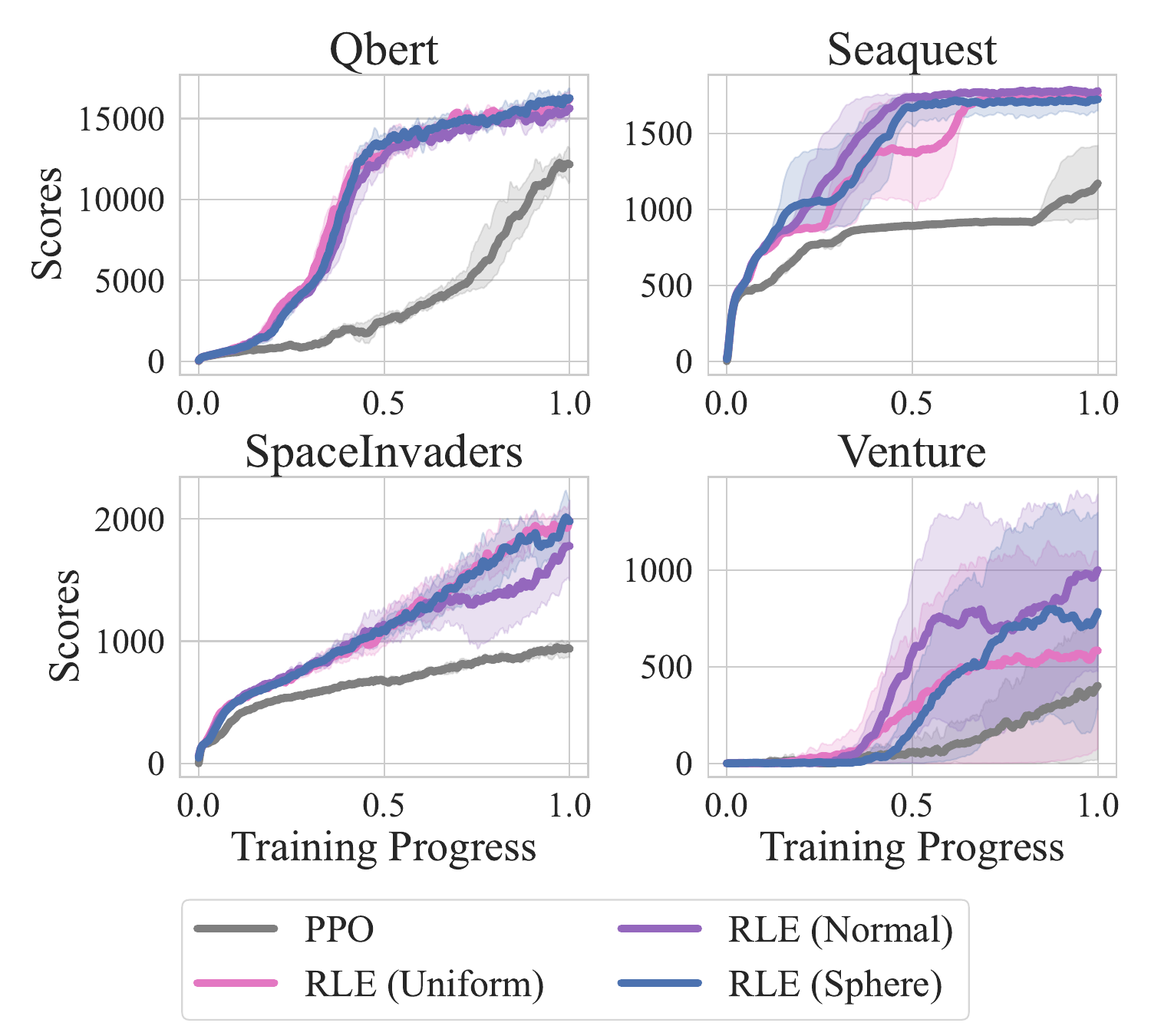}
    \caption{
 Performance of \RLE with varying latent vector distribution $P_{\boldsymbol{z}}$ (see Section~\ref{subsec:method:alg}), where \RLE (Sphere) is the one used in Section~\ref{subsec:exp:result}. The figure shows that \RLE with the three distributions can all outperform \PPO. This shows that \RLE is not sensitive to the choice of latent vector distribution.}
    \label{fig:ablation:dist}
\end{figure}

\begin{figure}[htb!]
    \centering
    \includegraphics[width=0.5\textwidth]{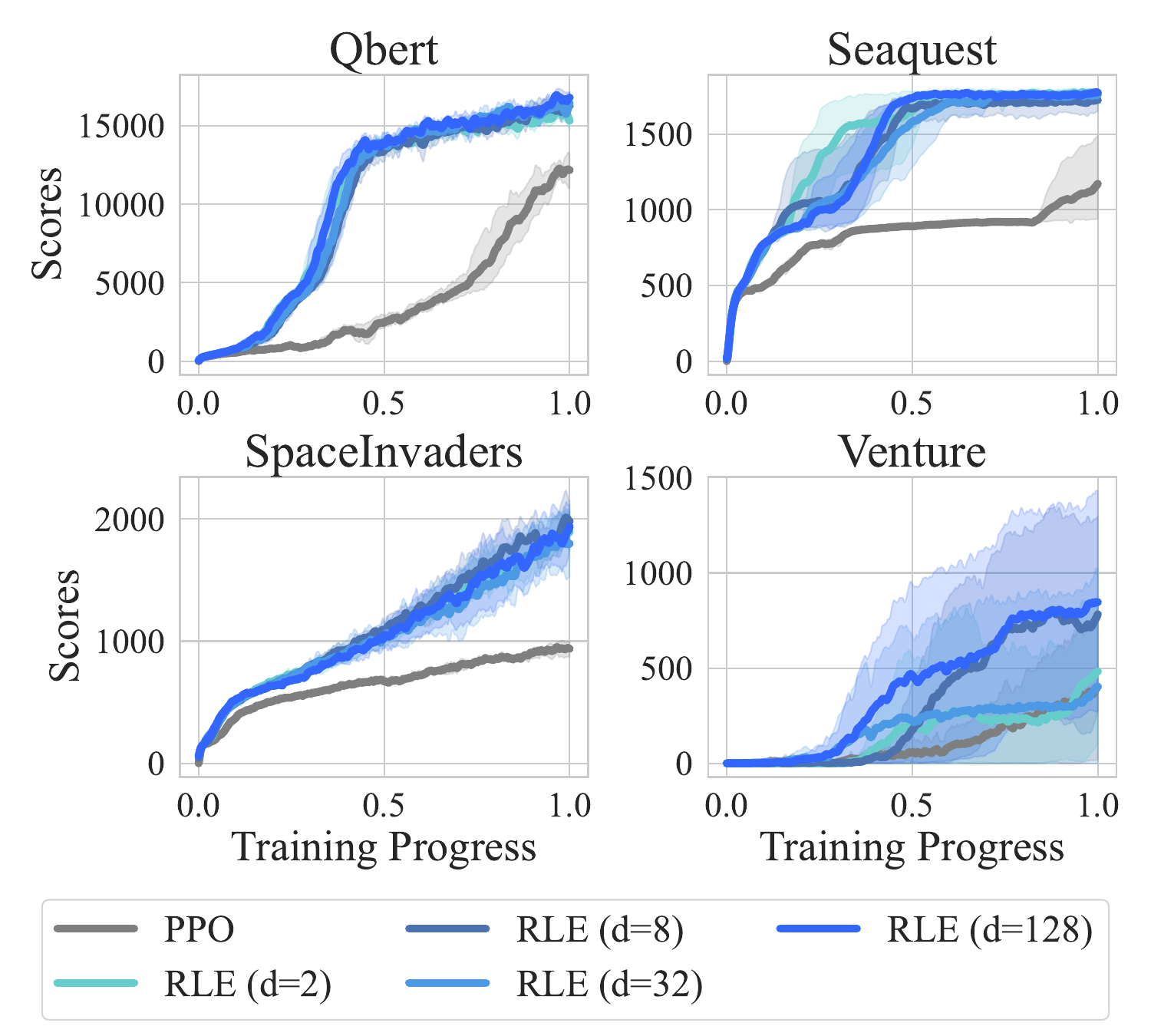}
    \caption{
 Performance of \RLE with varying latent vector dimension $d$ (see Section~\ref{subsec:method:alg}), where \RLE $(d=8)$ is the one used in Section~\ref{subsec:exp:result}. The figure shows that \RLE can outperform \PPO in all the four chosen dimensions. This shows that \RLE is not sensitive to the choice of latent vector dimension.}
    \label{fig:ablation:dim}
\end{figure}

\textbf{Effect of latent vector dimension.}
This study explores how robust is \RLE to different choices of the dimension $d$ of the latent vector \(\textbf{z}\). We trained \RLE for  $d \in \{2, 8, 32, 128\}$, where $d=8$ is the dimension used in the results presented in Section~\ref{subsec:exp:result}. The outcomes, depicted in Figure~\ref{fig:ablation:dim}, demonstrate that \RLE can surpass \PPO across all tested dimensions. Although slight performance variations exist between different $d$ values, these differences are subtle, suggesting that \RLE's performance is relatively insensitive to the choice of latent vector dimension $d$.

\begin{figure}[htb!]
    \centering
    \includegraphics[width=0.5\textwidth]{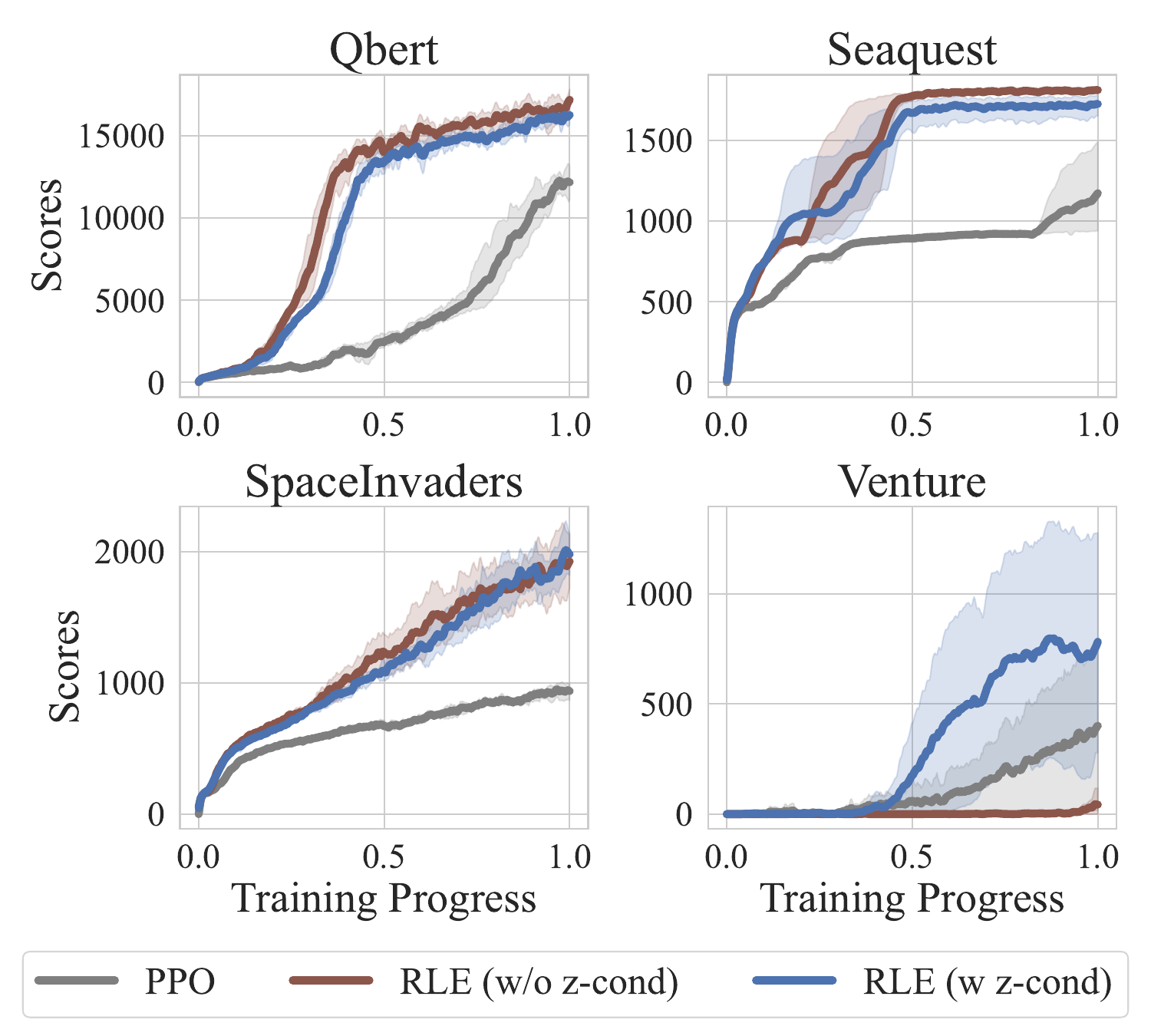}
    \caption{
 Performance of \RLE with (w z-cond) and without (w/o z-cond) latent vector condition in the policy and value networks (see Section~\ref{subsec:method:alg}). The figure displays that \RLE without latent vector condition suffers performance drop in \textsc{Venture}, a hard-exploration task with sparse rewards.}
    \label{fig:ablation:z_cond}
\end{figure}

\textbf{Latent vector conditioning.}
In Section~\ref{subsec:method:alg}, we emphasized the necessity for the policy to be conditioned on the latent vector to prevent randomized rewards $F(s, \boldsymbol{z})$ from being perceived as noise by both the policy and the value network. This design choice's importance is underscored by comparing \RLE models with and without latent-conditioned policy and value networks, as shown in Figure~\ref{fig:ablation:z_cond}. \RLE without latent vector conditioning exhibited a performance drop in the \textsc{Venture} task, a hard-exploration game with sparse rewards. We hypothesize that the absence of latent vector conditioning results in limited behavioral variability in the policy network, as its outputs remain unchanged by different latent vector samples. This limitation likely leads to failures in challenging exploration tasks that necessitate a broader diversity in trajectory generation. 

\paragraph{Choices of network architecture for the random reward network.} 
Since \RLE relies on neural networks to extract features for computing random rewards \( F(s, \boldsymbol{z}) \), it is important to examine how the choice of network architecture affects performance. We investigated the impact of different network architectures for extracting features \( \phi(s) \) on the computation of RLE rewards \( F(s, \boldsymbol{z}) \) in \textsc{IsaacGym} \citep{makoviychuk2021isaac}. We test only \textsc{IsaacGym} as \textsc{Atari} requires a different architecture to handle the image observation space, leaving no clear comparison with the architecture variations investigated here. We display the IQM of the normalized score in \Cref{fig:result:iqm_diff_arch} and POI over \PPO in \Cref{fig:result:poi_diff_arch}. In our original \textsc{IsaacGym} experiments, we used the value network's architecture for \RLE. In this ablation study, we tested a shallower neural network architecture. The results indicate that \RLE with a shallower network still performs well, suggesting that \RLE is not highly sensitive to the choice of network architecture.

\paragraph{White noise random rewards.}
In Section \ref{subsec:method:alg}, we emphasized the importance of ensuring that RLE's random reward function \( F \) is state-dependent.
We compare the state-dependent \RLE reward \( F(s, \boldsymbol{z}) \) with white noise rewards sampled from a normal distribution with zero mean and unit variance. We conduct this study in \textsc{IsaacGym} environments, with the results presented in \Cref{fig:result:isaacgym_noise_ablation}. The results demonstrate that white noise rewards significantly degrade performance, indicating that \RLE rewards are not merely white noise.

\section{Related Works} 
\label{sec:related}

Random reward prediction was used as an auxiliary task \citet{jaderberg2016reinforcement} for improving representation learning in prior works \citep{dabney2021value,lyle2021effect}.
A closely related work is \citet{ramesh2022exploring}, which employs a random general value function (GVF) \citep{sutton2011horde} for exploration by initializing a random reward function and using an ensemble of networks to predict policy-based random reward sums. The difference in prediction and the Monte Carlo estimate of random rewards, multiplied by prediction variance, enhances the agent's reward. Our work presents a distinct approach from \citet{ramesh2022exploring} both in terms of motivation and implementation. Contrary to \citet{ramesh2022exploring}, which aligns with previous studies \citep{burda2018exploration,pathak2017curiosity} by employing prediction errors as exploration bonuses, our \RLE algorithm directly trains the policy with random rewards, demonstrating superior performance. This finding underscores that \RLE provides a new angle to design exploration strategy beyond using prediction errors as exploration bonuses. Additionally, our RLE algorithm offers a more straightforward implementation by eliminating the need for ensemble training and Monte Carlo return estimation of random rewards. The detailed discussion on the relevant literature can be found in Appendix~\ref{app:related}.

\input{discussion}

\clearpage 

\subsection*{Acknowledgement} 
We thank members of the Improbable AI Lab for helpful discussions and feedback. We are grateful to MIT Supercloud and the Lincoln Laboratory Supercomputing Center for providing HPC resources. This research was supported in part by Hyundai Motor Company,  Quanta Computer Inc., MIT-IBM Watson AI Lab, an AWS MLRA research grant, ARO MURI under Grant Number W911NF-23-1-0277, ARO MURI under Grant Number W911NF-21-1-0328, and ONR MURI under Grant Number N00014-22-1-2740, NSF through award DMS-2031883, DOE through award DE-SC0022199, MIT UROP through the John Reed UROP Fund and the Paul E. Gray (1954) UROP Fund, and Simons Foundation. The views and conclusions contained in this document are those of the authors and should not be interpreted as representing the official policies, either expressed or implied, of the Army Research Office or the United States Air Force or the U.S. Government. The U.S. Government is authorized to reproduce and distribute reprints for Government purposes, notwithstanding any copyright notation herein. 

\subsection*{Author Contributions}
\begin{itemize}
    \item \textbf{Srinath Mahankali} ran initial experiments to investigate the benefit of random rewards that informed the eventual formulation of RLE, which he then compared against baseline methods on Atari and IsaacGym environments and helped with paper writing.
    \item \textbf{Zhang-Wei Hong} conceived the possibility of using random rewards for exploration. He was involved in research discussions, helped scale experiments, played a significant role in paper writing, and advised Srinath.
    \item \textbf{Ayush Sekhari} was involved in research discussions and helped set the overall formulation of RLE. He played a significant role in paper writing, and advised Srinath and Zhang-Wei. 
    \item \textbf{Alexander Rakhlin} was involved in research discussions and advising.
    \item \textbf{Pulkit Agrawal} was involved in research discussions, overall advising, paper writing, and positioning of the work.
\end{itemize}

\subsection*{Impact Statement} 
This paper presents work whose goal is to advance the field of Machine Learning. Our approach aims to accelerate RL in real-world domains, and depending on the domain to which it will be applied, we see many potential societal consequences of our work, however, none of them we feel must be specifically highlighted here.

\bibliography{main, references} 
\bibliographystyle{icml2024}

\appendix 
\onecolumn 

\section{Additional Related Works} 
\label{app:related}
\input{main_related_works.tex}

\clearpage

\input{experiment_details}
\input{appx_further_visualizations}

\input{detailed_results}

\end{document}

%% file: project_macros.tex
\newcommand{\RLE}{{\normalfont \textsc{RLE}}\xspace}  
\newcommand{\PPO}{{\normalfont \textsc{PPO}}\xspace} 
\newcommand{\RND}{{\normalfont \textsc{RND}}\xspace} 
\newcommand{\NoisyNet}{{\normalfont \textsc{NoisyNet}}\xspace} 
\newcommand{\FourRoom}{{\normalfont \textsc{FourRoom}}\xspace} 
\newcommand{\Atari}{{\normalfont \textsc{Atari}}\xspace} 
\newcommand{\IsaacGym}{{\normalfont \textsc{IsaacGym}}\xspace}  
\newcommand{\ShadowHand}{{\normalfont \textsc{ShadowHand}}\xspace}
\newcommand{\AllegroHand}{{\normalfont \textsc{AllegroHand}}\xspace}
\newcommand{\BallBalance}{{\normalfont \textsc{BallBalance}}\xspace}
\newcommand{\Humanoid}{{\normalfont \textsc{Humanoid}}\xspace}
\newcommand{\Ant}{{\normalfont \textsc{Ant}}\xspace}
\newcommand{\Cartpole}{{\normalfont \textsc{Cartpole}}\xspace}
\newcommand{\FrankaCabinet}{{\normalfont \textsc{FrankaCabinet}}\xspace}
\newcommand{\Anymal}{{\normalfont \textsc{Anymal}}\xspace}
\newcommand{\AnymalTerrain}{{\normalfont \textsc{AnymalTerrain}}\xspace}

%% file: new_intro.tex
Reinforcement learning (RL) \citep{sutton2018reinforcement} trains agents to maximize rewards through interactions with the environment. Since rewards can be delayed, focusing only on immediate rewards often leads to sub-optimal long-horizon strategies. Often, agents must sacrifice short-term rewards to discover higher rewards. Identifying actions that eventually result in higher rewards, known as the \textit{exploration} problem, is a core challenge in RL. 

Exploration is challenging because the current action's effect may often be revealed only after many interactions with the environment.  Exploration is well-studied~\citep{amin2021survey} and the major approaches can be broadly categorized into two types: (i) \textit{Noise-based exploration} (e.g., $\epsilon$-greedy, Boltzmann sampling, posterior sampling \citep{osband2016deep,osband2019deep,fortunato2017noisy,ishfaq2021randomized,ishfaq2023provable,plappert2017parameter}) and (ii) \textit{Bonus-based exploration}  \citep{bellemare2016unifying,pathak2017curiosity,burda2018large,oudeyer2009intrinsic,pathak2019self,hong2018diversity}.  
Unfortunately, neither family of approaches consistently outperforms the other when performance is measured across a range of tasks with either discrete \citep{chen2022redeeming} or continuous \citep{schwarke2023curiosity} action spaces. Unsurprisingly, the choice of exploration strategy for a new task is intimately tied to the task characteristics that are difficult to determine in advance. Therefore, the current common practice is finding the best exploration scheme using the trial-and-error process of trying different strategies.   

Noise-based exploration typically involves perturbing the policy's parameters (such as the weights of the policy network \citep{fortunato2017noisy,plappert2017parameter}), or the action output of the policy (e.g.~\(\varepsilon\)-greedy). The added noise prevents the agent from generating the same trajectory repeatedly, thereby encouraging the exploration of different trajectories. 
Noise-based exploration is the go-to exploration scheme in deep RL due to its simplicity of implementation. However, such strategies are less effective in tasks requiring deep exploration than bonus-based exploration strategies \citep{osband2016deep}. One possible reason revealed by our experiments (see Section~\ref{subsec:exp:illustrative}) is that commonly used perturbation strategies only affect the policy locally and, therefore, do not explore states far from the initial states. %

Bonus-based strategies augment the task reward with an incentive (or a bonus) (e.g., prediction error \citep{pathak2017curiosity}, visitation count \citep{bellemare2016unifying}, information gain \citep{houthooft2016vime}) that encourages the agent to explore far away states and thereby achieve deep exploration. 
Unfortunately, computing the bonus requires training an additional deep neural network. Furthermore, while bonus-based exploration outperforms noise-based exploration in a few hard-exploration tasks, when average performance is measured across a range of tasks (i.e., both easy and hard exploration problems), both strategies perform similarly~\cite{taiga2019benchmarking, chen2022redeeming}.

Instead, we hypothesize that it is easier to explore by training the agent to achieve a variety of diverse goals, i.e., exploring the space of goals~\cite{smith2005development,forestier2022intrinsically}. We build upon this intuition and introduce an exploration strategy where instead of injecting noise or adding bonuses to encourage exploration, we train the agent to achieve various goals from a goal space \(\mathcal{Z}\).  Our key intuition is that if the goals are designed such that different goals incentivize the agent to explore diverse parts of the environment, then training an agent to achieve these goals will lead it to explore a variety of states, eventually leading it to find and achieve high task rewards. 
While goal-based exploration has been well studied~\cite{nair2018visual,ecoffet2019go,ecoffet2021first,torne2023breadcrumbs}, representing and selecting "the right goals" is challenging in practice, as they may be dependent on the underlying task in complex ways that the learner cannot anticipate in advance.

To circumvent the issue of finding the right goals to target, we propose \textit{Random Latent Exploration (RLE)}, where the agent's policy is conditioned on random vectors sampled, $z \sim \mathcal{Z}$, from a fixed distribution as goals. Each goal ($z$) defines an exploration bonus at every state, and different goals define different bonuses. In particular, the random vectors act as latent goals, each creating different reward functions that encourage the agent to reach different parts of the environment. By sampling enough random vectors during training, the agent is trained to pursue many different goals, thus resulting in deep exploration. Our experiments show that RLE leads to significantly more diverse and deeper trajectories than traditional noise-based methods, without the need for complex bonus computation. This makes RLE both easy to implement and effective in practice. 

To show the effectiveness of our approach, we evaluated RLE in \Atari---a popular discrete action space deep RL benchmark  \citep{bellemare2013arcade}, and \IsaacGym---a popular continuous control deep RL benchmark \citep{makoviychuk2021isaac}, each consisting of many different tasks with varying degrees of exploration difficulty. 
We implement our method on top of the popular RL algorithm, Proximal Policy Optimization (PPO) \citep{schulman2017proximal} and compare it with PPO in other exploration strategies. 
Our experimental results demonstrate that \RLE improves over standard PPO in \Atari and \IsaacGym. Furthermore, RLE also exhibits a higher aggregated score across all tasks in \Atari than other exploration methods, including RND~\cite{burda2018exploration} and randomized value function strategies \citep{fortunato2017noisy}. Importantly, these improvements were obtained by simply adding RLE on the top of the base \PPO implementation while only changing  the learning rate (and discount rate for \Atari). We use the same hyperparameters across all 57 \Atari games and the same hyperparameters across all 9 \IsaacGym tasks, highlighting the generality of our approach as a plug-in utility.

%% file: discussion.tex
\section{Discussion and Conclusion} 
In this paper, we proposed a new exploration method called \RLE that is straightforward to implement in practice, and effective in challenging deep RL benchmarks like \Atari. We conclude with discussions and future work directions:  

\textbf{Simple plug-in for deep RL algorithms.} \RLE simply requires adding randomized rewards to the rewards received by the agent and augmenting the agent's input with additional latent variables that correlate to these randomized rewards. As a result, \RLE is agnostic to the base RL algorithm and can be integrated with any RL algorithm. Given its simplicity, generality, and the overall performance improvement it provides, we recommend using \RLE as the default exploration strategy in deep RL implementations.

\input{posterior_sampling.tex}

\textbf{Benefits from parallelization.} Note that, by design, our algorithm samples independent \(\boldsymbol{z}\) in every round and can thus benefit from parallelization by running the algorithm on multiple copies of the same environment (when possible, e.g. using a simulator). Since different \(\boldsymbol{z}\) produce diverse trajectories (see \Cref{fig:Trajectory_Diversity1} or \Cref{app:fig:full_rle_trajectory_plots} for illustrations), multiple parallel copies of the same agent will simply produce more diverse data which would accelerate exploration.

\textbf{On the inductive bias of \(\boldsymbol{\phi}\).}  \RLE is modular as one can choose any feature extractor $\phi(s)$  e.g.  Transformer networks~\cite{vaswani2017attention}, MLPs, or even nonparametric models such as kernels. In our \Atari experiments we use a CNN for $\phi$, but it would be interesting to explore how other choices of $\phi$ affect the diversity of the induced reward functions, and hence the generated trajectories. 

 \paragraph{\(z\)-sampling.} 
 At every timestep in Algorithm \ref{alg:alg} the latent variable  \(\boldsymbol{z}\) is sampled independently from the fixed distribution \(P_z\) which is chosen at initialization. However, it is also intuitive to expect that \(P_z\) should change as we learn more about the underlying environment. Looking forward it would be interesting to explore algorithms that change \(P_z\) while training, e.g. to sample more from the set of latent variables which have not been explored yet or for which the corresponding policies \(\pi_z\) have historically performed well in the given environment.

\textbf{Other limitations.}  
Currently, we limit our study to on-policy algorithms. Looking forward it would be interesting to extend \RLE to off-policy algorithms such as DQN~\citep{mnih2015human} and SAC~\cite{haarnoja2018soft}; A practical way to do so would be to condition the $Q$-function on $\boldsymbol{z}$ in addition to its usual inputs. Separately, an important direction for future work is to explore the method in more continuous control and real-world robotics domains. While it is clear that our approach scales to high-dimensional state spaces in \Atari and continuous control tasks in \(\IsaacGym\), it would also be interesting to see how it would generalize for real-world RL applications, e.g.~in robotics. Finally, while \RLE leads to performance improvements in general, we note that it does not improve performance in the famous hard-exploration game \textsc{Montezuma's Revenge}. Thus, in the future, it would be interesting to extend \RLE to environments and tasks with even sparser rewards.

\clearpage

%% file: posterior_sampling.tex
\textbf{Connection to posterior sampling.} At a high level, while \RLE seems similar to the posterior sampling-based approaches  \citep{thompson1933likelihood, thompson1935theory, russo2018tutorial} in the sense that both utilize randomization for exploration, there are important differences: Firstly, the two methods explore via different mechanisms. Posterior sampling randomizes over different models of the environment, whereas \RLE perturbs the reward function using random rewards. Secondly, the sampling distribution \(P_z\) is fixed throughout learning in \RLE, whereas the posterior distribution in posterior sampling changes with time and needs to be computed for every round (which is often challenging in practice). Thirdly, in posterior sampling, the posterior will eventually concentrate around the true model, and thus the algorithm will execute the optimal policy for the underlying environment. Whereas, in \RLE, since the task rewards are constant throughout learning whereas random rewards change, in the later stage of the learning, the trained policy \(\pi_z\) should focus on optimizing just the task rewards, and we believe that the random rewards will simply act as a regularization.

%% file: main_related_works.tex
There are two main approaches in RL  for exploration, (a) Randomize the agent by injecting noise thus inducing diverse behavior, and (b) Provide explicit reward bonuses that incentivize the agent to go to novel states. \RLE, bridging the two approaches, injects noise into the agent by adding random latent reward bonuses during training. 

Randomness is the key tool in many exploration strategies in RL.  Perhaps the most popular example is  $\epsilon$-greedy or Boltzmann sampling-based exploration~\cite{mnih2013playing, dann2022guarantees, cesa2017boltzmann, eysenbach2018diversity},  which explores by playing random actions. Entropy regularization~\cite{williams1991function,mnih2016asynchronous}, and MaxEnt RL~\cite{haarnoja2018soft,eysenbach2021maximum, garg2023extreme,hazan2019provably} are other instances of exploration algorithms that utilize randomness, as they explicitly bias towards learning policies that have a high entropy. Another exploration approach is to directly inject noise into the parameters of the policy or value networks, e.g. in off-policy methods ~\cite{fortunato2017noisy}, RLHF with linear MDPs \citep{wu2023making}, online RL in tabular  \citep{osband2016generalization}, and linear MDPs \citep{zanette2020frequentist}. Another line of work includes using Thompson sampling \citep{thompson1933likelihood, thompson1935theory, russo2018tutorial}, or posterior sampling for exploration \citep{osband2013more, gopalan2015thompson, kveton2021meta, zhang2022feel}, which maintains a posterior distribution over the ground truth model and relies on the uncertainty in the posterior distribution for exploration. Posterior sampling, however, is intractable in practical RL settings due to the need to sample from the extremely complex posterior distribution; Various empirical approaches aim to sample from an approximate posterior instead \citep{li2021hyperdqn,dwaracherla2020hypermodels}, but are unfortunately memory intensive. We note that \RLE is different from these other works as it explores by adding random rewards instead of randomizing over the actions, policies, or models of the environment. 

Exploration by adding explicit reward bonuses is also well studied in both theoretical and applied RL literature. A popular technique is to add novelty-based exploration bonuses that are constructed using prediction errors in transition dynamics ~\cite{pathak2017curiosity,pathak2019self,ramesh2022exploring} or outputs of randomly initialized target network \RND~\citep{burda2018exploration}, etc. Other approaches construct reward bonuses using upper confidence bounds on the uncertainty estimates for the underlying model \citep{auer2008near, vaswani2019old}, using discriminatively trained exemplar models to estimate novelty \citep{fu2017ex2}, or using elliptical potentials when the MDP has a linear parameterization \citep{jin2020provably, zhang2022making, agarwal2020pc}.  {Unfortunately, these methods often introduce additional components into the learning setups, e.g. additional neural networks for generating bonuses and the associated hyperparameters, which can make learning unstable. In contrast, \RLE is much simpler to deploy as it adds random reward functions that are computed using features from the policy network.} 

\RLE closely resembles the idea of  \textit{Follow The Perturbed Leader} (FTPL) developed in RL theory literature \citep{kveton2019garbage, kveton2019perturbed, kveton2020randomized, rakhlin2016bistro, dai2022follow}. FTPL-based methods explore by adding carefully designed perturbations to the reward function that can guarantee optimism; since the perturbations are closely tied to the underlying modeling assumptions, FTPL-based methods are currently limited to restricted settings like linear bandits, tabular MDPs, etc. In contrast, \RLE simply adds a random reward function sampled from a fixed distribution \(P_z\), and is thus applicable in more practical RL settings. Another major difference is that our method also utilizes \(z\)-conditioned policies and value functions, and thus the randomness is shared amongst the reward function, policy network and value network. 

Finally, there is a long line of work in the theoretical RL literature on developing exploration algorithms that can optimally exploit the structure of the underlying task in model-based RL \citep{ayoub2020model, sun2019model, foster2021statistical}, model-free RL \citep{jin2021bellman, du2021bilinear} and agnostic RL setting \citep{jia2023agnostic},  
however, the focus in these works is on statistical efficiency and the provided approaches are not computationally tractable.

%% file: experiment_details.tex
\section{Experimental Implementation Details} \label{app:impl} 
In this section, we provide the hyperparameters and implementation details of our algorithm (\RLE) along with the baseline methods (\PPO, \RND, \NoisyNet) for the \FourRoom and \Atari Environments. We also provide hyperparameters and implementation details for all \IsaacGym experiments.

\subsection{\RLE Pseudocode} 
Below, we provide the pseudocode for \RLE in \Cref{app:alg:alg}.

\begin{algorithm}[H]
\begin{algorithmic}[1]
    \STATE \textbf{Input:} Latent distribution $P_{\boldsymbol{z}}, N$ parallel workers, $T$ steps per update, $S$ steps per sampling, feature network update rate $\tau$ 
    \STATE Randomly initialize a feature network $\phi$ with the same backbone architecture as the policy and value networks
    \STATE Initialize running mean $\boldsymbol{\mu} = \mathbf{0}$ and standard deviation $\boldsymbol{\sigma} = \mathbf{1}$ estimates of $\phi(s)$ over the state space
    \STATE Sample an initial latent vector for each parallel worker: $\boldsymbol{z} \sim P_{\boldsymbol{z}}$     
    \REPEAT  
        \STATE Sample initial state \(s_0\). 
        \FOR {$t = 0, \dots, T$} 
            \STATE Take action $a_t \sim \pi(.| s_t, \boldsymbol{z})$ and transition to $s_{t+1}$ 
            \STATE Compute feature $f(s_{t+1}) = (\phi(s_{t+1}) - \boldsymbol{\mu})/\boldsymbol{\sigma}$
            \STATE Compute random reward: $F(s_{t+1}, \boldsymbol{z}) = \frac{f(s_{t+1})}{\|f(s_{t+1})\|} \cdot z$
            \STATE  Receive  reward: $r_t = R(s_t, a_t) + F(s_{t+1}, \boldsymbol{z})$
            \FOR {$i = 0, 1, \dots, N-1$}
                \IF {{worker $i$ terminated} \OR {$S$ timesteps passed without resampling}}
                     \STATE Resample sample $\boldsymbol{z} \sim P_{\boldsymbol{z}}$ for worker $i$
                \ENDIF
            \ENDFOR
        \ENDFOR
        \STATE Update policy network $\pi$ and value network $V^\pi$ with the collected trajectory $(\boldsymbol{z}, s_0, a_0, r_0, s_1, \cdots, s_{T})$
        \STATE Update feature network $\phi$ using the value network's parameters: $\phi \leftarrow \tau \cdot \pi + (1 - \tau) \cdot \phi$
        \STATE Update $\boldsymbol{\mu}$ and $\boldsymbol{\sigma}$ using the batch of collected experience. 
    \UNTIL convergence       
\end{algorithmic}
    \caption{Detailed Pseudocode for Random Latent Exploration (\RLE)} 
    \label{app:alg:alg}
\end{algorithm}

\subsection{\textsc{FourRoom} Environment} 
\label{app:four_room_details}
We provide the  hyperparameters used for experiments in the \FourRoom environment in Table \ref{tabl:hyperparameters_fourroom}. 
\begin{table}[ht]
    \begin{center}
        \begin{tabular}{p{6.5cm}|p{4.5cm}}
        \toprule
        Parameter & Value \\

        \midrule
        \PPO \\
        \hspace{15pt} Total Timesteps & $2,500,000$ \\
        \hspace{15pt} Optimizer & Adam \\
        \hspace{15pt} Learning Rate & $0.001$ \\
        \hspace{15pt} Adam Epsilon & $0.00001$ \\
        \hspace{15pt} Parallel Workers & $32$ \\
        \hspace{15pt} Steps per Batch & $128$ \\
        \hspace{15pt} Discount Rate & $0.99$ \\
        \hspace{15pt} Generalized Advantage Estimation $\lambda$ & $0.95$ \\
        \hspace{15pt} Minibatches per Epoch & $4$ \\
        \hspace{15pt} Epochs per Training Step & $4$ \\
        \hspace{15pt} Clipping Coefficient & $0.2$ \\
        \hspace{15pt} Entropy Loss Weight & $0.01$ \\
        \hspace{15pt} Discount Rate & $0.99$ \\
        \hspace{15pt} Value Loss Weight & $0.5$ \\
        \hspace{15pt} Gradient Norm Bound & $0.5$ \\
        \hspace{15pt} Use Advantage Normalization & True \\
        \hspace{15pt} Use Clipped Value Loss & True \\
        \hspace{15pt} Policy Network Architecture & MLP (64,64,4) \\
        \hspace{15pt} Value Network Architectures & MLP (64,64,1) \\
        \hspace{15pt} Network Activation & Tanh \\
        \midrule
        \NoisyNet \\
        \hspace{15pt} Initial $\sigma$ & $0.017$ \\
        \midrule 
        \RND \\
        \hspace{15pt} Intrinsic Reward Coefficient & $1.0$ \\
        \hspace{15pt} Drop Probability & $0.25$ \\
        \hspace{15pt} Predictor Network Architecture & MLP (256, 256, 256, 256, 256) \\
        \hspace{15pt} Target Network Architecture & MLP (64,256) \\
        \hspace{15pt} Network Activation & ReLU \\
        \midrule 
        \RLE & \\
        \hspace{15pt} Intrinsic Reward Coefficient & $0.1$ \\
        \hspace{15pt} Latent Vector Dimension & $4$ \\
        \hspace{15pt} Feature Network Architecture & MLP (64,64,64,4) \\ 
        \hspace{15pt} Network Activation & ReLU \\
        \bottomrule
        \end{tabular}
        \label{app:table:grid_hyper}
    \end{center}
    \caption{Hyperparameters and network architectures for \FourRoom experiments.} 
    \label{tabl:hyperparameters_fourroom}
\end{table}

\subsubsection{\textsc{RLE} Implementation in \textsc{FourRoom} Enviroment} 
In our implementation of \RLE for \FourRoom environment, we ensure that the random reward functions $F(s,z)$ take values in $[-1,1]$.To compute the reward given a state $s$ and latent variable $\boldsymbol{z}$, we normalize the output of $\phi(s)$ to have unit norm. Specifically, we define the reward as: 
\begin{equation*}
    F(s,\boldsymbol{z}) = \frac{\phi(s)}{\|\phi(s)\|} \cdot z,
\end{equation*}
where $\phi$ is the randomly initialized feature network that transforms the state $s$ to a vector with the same dimension as $\boldsymbol{z}$. In the \FourRoom environment, we sample $\boldsymbol{z}$ from the unit sphere at every training step, which occurs every $128$ timesteps. We perform the sampling independently for each of the $32$ parallel workers. 

\subsection{\textsc{Atari}}

We display the hyperparameters used for experiments in \Atari games in Table \ref{tabl:atari_hyperparameters}. For \PPO and \RND, we use the default hyperparameters based on the \url{cleanrl} codebase~\citep{huang2022cleanrl}, which were tuned for \Atari games. For \NoisyNet, we use the same hyperparameters as \PPO with the exception of the entropy loss weight, which is set to $0$ as recommended by \citep{fortunato2017noisy}. We give a detailed description of the \Atari implementation of \RLE below.

\begin{table}[ht!]
    \begin{center}
        \begin{tabular}{p{6.5cm}|p{4.5cm}}
        \toprule
        Parameter & Value \\

        \midrule
        \PPO \\
        \hspace{15pt} Total Timesteps & $40,000,000$ \\
        \hspace{15pt} Optimizer & Adam \\
        \hspace{15pt} Learning Rate & $0.0001$ \\
        \hspace{15pt} Adam Epsilon & $0.00001$ \\
        \hspace{15pt} Parallel Workers & $128$ \\
        \hspace{15pt} Steps per Batch & $128$ \\
        \hspace{15pt} Discount Rate & $0.99$ \\
        \hspace{15pt} Generalized Advantage Estimation $\lambda$ & $0.95$ \\
        \hspace{15pt} Minibatches per Epoch & $4$ \\
        \hspace{15pt} Epochs per Training Step & $4$ \\
        \hspace{15pt} Clipping Coefficient & $0.1$ \\
        \hspace{15pt} Entropy Loss Weight & $0.01$ \\
        \hspace{15pt} Discount Rate & $0.99$ \\
        \hspace{15pt} Value Loss Weight & $0.5$ \\
        \hspace{15pt} Gradient Norm Bound & $0.5$ \\
        \hspace{15pt} Use Advantage Normalization & True \\
        \hspace{15pt} Use Clipped Value Loss & True \\
        \hspace{15pt} Policy Network Architecture & CNN + MLP (256,448,448,18) \\
        \hspace{15pt} Value Network Architectures & CNN + MLP (256,448,448,1) \\
        \hspace{15pt} Network Activation & ReLU \\
        \midrule
        \NoisyNet \\
        \hspace{15pt} Initial $\sigma$ & $0.017$ \\
        \hspace{15pt} Entropy Loss Weight & $0$ \\
        \midrule 
        \RND \\
        \hspace{15pt} Intrinsic Reward Coefficient & $1.0$ \\
        \hspace{15pt} Extrinsic Reward Coefficient & $2.0$ \\
        \hspace{15pt} Update Proportion & $0.25$ \\
        \hspace{15pt} Observation Normalization Iterations & $50$ \\
        \hspace{15pt} Discount Rate & $0.999$ \\
        \hspace{15pt} Entropy Loss Weight & $0.001$ \\
        \hspace{15pt} Intrinsic Discount Rate & $0.99$ \\
        \hspace{15pt} Predictor Network Architecture & CNN + MLP (512,512,512) \\
        \hspace{15pt} Target Network Architecture & CNN + MLP (512) \\
        \hspace{15pt} Network Activation & LeakyReLU \\
        \midrule 
        \RLE & \\
        \hspace{15pt} Intrinsic Reward Coefficient & $0.01$ \\
        \hspace{15pt} Latent Vector Dimension & $16$ \\
        \hspace{15pt} Latent Vector Resample Frequency & $1280$ \\
        \hspace{15pt} Learning Rate & $0.0003$ \\
        \hspace{15pt} Discount Rate & $0.999$ \\
        \hspace{15pt} Intrinsic Discount Rate & $0.99$ \\
        \hspace{15pt} Feature Network Update Rate $\tau$ & $0.005$ \\
        \hspace{15pt} Feature Network Architecture & CNN + MLP (256,448, 16) \\ 
        \hspace{15pt} Network Activation & ReLU \\
        \bottomrule
        \end{tabular}
        \label{app:table:atari_hyper}
    \end{center}
    \caption{ Hyperparameters and network architectures for \Atari experiments.}
    \label{tabl:atari_hyperparameters}
\end{table}

\subsubsection{\textsc{RLE} Implementation Details in \textsc{Atari}}
\label{subsubsec:rle_imp_details_atari}
\paragraph{Feature network architecture and update.} 
We start with a randomly initialized neural network $\phi$ which takes a state $s$ as input and outputs a vector in $\R^d$, which has the same dimension as $\boldsymbol{z}$. In our implementation, $\phi$ contains a CNN backbone with an identical architecture to the (shared) policy and value backbone, along with a final linear layer on top to convert it to a low dimension $\R^d$. In our implementation, we choose $d = 8$. To update $\phi$, we follow the rule:
\begin{equation*}
    \mathrm{CNN}_\phi \leftarrow \tau \cdot \mathrm{CNN}_{V} + (1 - \tau) \cdot \mathrm{CNN}_\phi
\end{equation*}
for a small value of $\tau$, and we choose $\tau = 0.005$ for our experiments. This network update is inspired by the target network update in DQN~\citep{mnih2015human} and does not require any gradient steps.

\paragraph{Computation of random reward bonus.}
When the agent experiences a transition $(s,a,s')$, we obtain random reward bonus from $\phi$ as follows: Obtain the low-dimensional vector output  $\phi(s')$. We standardize the output of $\phi(s')$ using a running mean and standard deviation estimate so that the output is a normal distribution on $\R^d$.  Meanwhile, sample a vector $\boldsymbol{z}\sim\mathcal{S}^{d-1}$, and compute the following value
\begin{equation*}
    F(s', \boldsymbol{z})= \frac{\phi(s') \cdot \boldsymbol{z}}{\|\phi(s')\|}.
\end{equation*}

\paragraph{Policy input.}
The policy observes the observation returned by the environment, which is $4$ stacked grayscale frames of dimension $(84,84)$. In addition, the policy observes $\boldsymbol{z}$ as well as the random reward $F(s_t, \boldsymbol{z})$ from the previous time step. 

\paragraph{Resampling of the latent variable \(\boldsymbol{z}\).} 
In our Atari experiments, there are 128 parallel workers. We sample $\boldsymbol{z}$ independently across all workers from the $d$-dimensional unit sphere $\mathcal{S}^{d-1}$, and resample upon either of the following signals:
\begin{enumerate}
    \item An environment has reached the `DONE' flag, or  
    \item An environment has survived with this $\boldsymbol{z}$ for $1280$ time steps.
\end{enumerate} 

\paragraph{Policy training.}
We use PPO with the augmented observation space train on the combined reward as usual. As we resample $\boldsymbol{z}$ during an episode, we also treat the problem of maximizing randomized reward as episodic. Specifically, we set the  `done' signal to True whenever we resample $\boldsymbol{z}$. Thus, we do not use returns from future $\boldsymbol{z}$ within the same episode to estimate the return under the current $\boldsymbol{z}$. 

\subsubsection{Evaluation Details in Atari}\label{app:atari_eval_details}

\paragraph{Human normalized score} To compute aggregate performance, we first compute the human normalized score for each seed in each environment as $\frac{\mathrm{Agent}_{\mathrm{score}} - \mathrm{Random}_{\mathrm{score}}}{\mathrm{Human}_{\mathrm{score}} - \mathrm{Random}_{\mathrm{score}}}$. After this, we compute the IQM to measure aggregate performance as recommended in~\cite{agarwal2021deep} as it is robust to outliers.

\paragraph{Capped human normalized score} We use the capped human normalized score (CHNS)~\citep{badia2020agent57} to measure the aggregate performance of \RLE and baselines in \Cref{fig:atari_chns_mean}. To compute the CHNS, we first compute the human normalized score (HNS) of the agent, as done in~\citep{badia2020agent57}, as $\frac{\mathrm{Agent}_{\mathrm{score}} - \mathrm{Random}_{\mathrm{score}}}{\mathrm{Human}_{\mathrm{score}} - \mathrm{Random}_{\mathrm{score}}}$, after which it is clipped to be between $0$ and $1$. In addition to aggregate metrics, we provide individual mean scores of all methods in all 57 games in \Cref{tab:atari_results} along with the corresponding learning curves in \Cref{fig:atari_learning_curves}.

\paragraph{Probability of improvement} We use the probability of improvement (POI), recommended in~\cite{agarwal2021deep}, to measure the relative performance between algorithms across all 57 \Atari games.

\paragraph{Bootstrapped confidence intervals} We use the bootstrapping method~\citep{diciccio1996bootstrap, agarwal2021deep} to estimate the confidence intervals for all aggregated metrics we report, and mean performance for an algorithm in one environment.

\subsection{\IsaacGym}\label{subsec:isaacgym_details}

We display the hyperparameters used for experiments in IsaacGym in \Cref{tabl:isaacgym_hyperparameters}. For \PPO, we use the default hyperparameters recommended by the \url{cleanrl} codebase~\citep{huang2022cleanrl}, which were tuned for IsaacGym tasks and can vary across tasks (specifically, using different hyperparameters for the \ShadowHand and \AllegroHand tasks). For \RLE, we use the same hyperparameters for each task. We also display the environment-specific hyperparameters in \Cref{table:isaacgym_environment_params}, which are shared for each training algorithm we consider in our experiments.

\begin{table}[ht!]
    \begin{center}
        \begin{tabular}{p{6.5cm}|p{4.5cm}}
        \toprule
        Parameter & Value \\

        \midrule
        \PPO \\
        \hspace{15pt} Optimizer & Adam \\
        \hspace{15pt} Learning Rate & $0.0026$ \\
        \hspace{15pt} Adam Epsilon & $0.00001$ \\
        \hspace{15pt} Steps per Batch & $16$ \\
        \hspace{15pt} Discount Rate & $0.99$ \\
        \hspace{15pt} Generalized Advantage Estimation $\lambda$ & $0.95$ \\
        \hspace{15pt} Minibatches per Epoch & $2$ \\
        \hspace{15pt} Epochs per Training Step & $4$ \\
        \hspace{15pt} Clipping Coefficient & $0.2$ \\
        \hspace{15pt} Entropy Loss Weight & $0.0$ \\
        \hspace{15pt} Discount Rate & $0.99$ \\
        \hspace{15pt} Value Loss Weight & $2.0$ \\
        \hspace{15pt} Gradient Norm Bound & $1.0$ \\
        \hspace{15pt} Use Advantage Normalization & True \\
        \hspace{15pt} Use Clipped Value Loss & False \\
        \hspace{15pt} Policy Network Architecture & MLP (256,256,256) \\
        \hspace{15pt} Value Network Architecture & MLP (256,256,256,1) \\
        \hspace{15pt} Network Activation & Tanh \\
        \hspace{15pt} Reward Scale & $1.0$ \\
        \midrule
        \RND & \\
        \hspace{15pt} Intrinsic Value Loss Weight & $2.0$ \\
        \hspace{15pt} Intrinsic Reward Coefficient & $0.5$ \\
        \hspace{15pt} Update Proportion & $0.0625$ \\
        \hspace{15pt} Observation Normalization Iterations & $50$ \\
        \hspace{15pt} Predictor Network Architecture & MLP (256, 256, 256, 256) \\
        \hspace{15pt} Target Network Architecture & MLP (64, 64, 256) \\
        \hspace{15pt} Network Activation & ReLU \\
        \midrule
        \RLE & \\
        \hspace{15pt} Intrinsic Value Loss Weight & $0.5$ \\
        \hspace{15pt} Intrinsic Reward Coefficient & $0.01$ \\
        \hspace{15pt} Latent Vector Dimension & $32$ \\
        \hspace{15pt} Latent Vector Resample Frequency & $16$ \\
        \hspace{15pt} Learning Rate & $0.001$ \\
        \hspace{15pt} Feature Network Update Rate $\tau$ & $0.005$ \\
        \hspace{15pt} Policy Network Architecture & MLP (256,256,256) \\
        \hspace{15pt} Value Network Architecture & MLP (512,512,256,256,1) \\
        \hspace{15pt} Feature Network Architecture & MLP (512,512,256) \\ 
        \hspace{15pt} Network Activation & Tanh \\
        \midrule 
        \PPO (\AllegroHand and \ShadowHand) \\
        \hspace{15pt} Steps per Batch & $8$ \\
        \hspace{15pt} Minibatches per Epoch & $4$ \\
        \hspace{15pt} Epochs per Training Step & $5$ \\
        \hspace{15pt} Reward Scale & $0.01$ \\
        \bottomrule
        \end{tabular}
        \label{app:table:isaac_hyper}
    \end{center}
    \caption{Hyperparameters and network architectures for IsaacGym experiments. The number of training steps and parallel workers depends on the environment, but are shared across different methods.}
    \label{tabl:isaacgym_hyperparameters}
\end{table}

\begin{table}[ht!]
    \begin{center}
        \begin{tabular}{p{6.5cm}|p{4.5cm}}
        \toprule
        Parameter & Value \\
        
        \midrule
        \AllegroHand \\
        \hspace{15pt} Number of Timesteps & $600,000,000$ \\
        \hspace{15pt} Number of Parallel Environments & $8,192$ \\
        \midrule
        \ShadowHand \\
        \hspace{15pt} Number of Timesteps & $600,000,000$ \\
        \hspace{15pt} Number of Parallel Environments & $8,192$ \\
        \midrule
        \BallBalance \\
        \hspace{15pt} Number of Timesteps & $200,000,000$ \\
        \hspace{15pt} Number of Parallel Environments & $4,096$ \\
        \midrule 
        \Humanoid \\
        \hspace{15pt} Number of Timesteps & $200,000,000$ \\
        \hspace{15pt} Number of Parallel Environments & $4,096$ \\
        \midrule
        \Ant \\
        \hspace{15pt} Number of Timesteps & $100,000,000$ \\
        \hspace{15pt} Number of Parallel Environments & $4,096$ \\
        \midrule
        \Cartpole \\
        \hspace{15pt} Number of Timesteps & $100,000,000$ \\
        \hspace{15pt} Number of Parallel Environments & $4,096$ \\
        \midrule
        \FrankaCabinet \\
        \hspace{15pt} Number of Timesteps & $100,000,000$ \\
        \hspace{15pt} Number of Parallel Environments & $4,096$ \\
        \midrule
        \Anymal \\
        \hspace{15pt} Number of Timesteps & $100,000,000$ \\
        \hspace{15pt} Number of Parallel Environments & $4,096$ \\
        \midrule
        \AnymalTerrain \\
        \hspace{15pt} Number of Timesteps & $100,000,000$ \\
        \hspace{15pt} Number of Parallel Environments & $4,096$ \\
        \bottomrule
        \end{tabular}
        \caption{Environment-specific parameters and their values. These parameters are shared across all algorithms.}
        \label{table:isaacgym_environment_params}
    \end{center}
\end{table}

\subsubsection{\textsc{RLE} Implementation Details in \textsc{IsaacGym}}

\paragraph{Feature network architecture and update} Similar to our implementation of RLE in the \Atari domain, we start with a randomly initialized neural network $\phi$ that has an MLP backbone with the same architecture as the backbone of the value function. We update the backbone paramaters using the same slow moving average as in \Cref{subsubsec:rle_imp_details_atari} with $\tau = 0.005$:
\begin{equation*}
    \mathrm{MLP}_\phi \leftarrow \tau \cdot \mathrm{MLP}_{V} + (1 - \tau) \cdot \mathrm{MLP}_\phi.
\end{equation*}

\paragraph{Computation of random reward bonus} We standardize the output of $\phi(s')$ using a running mean and standard deviation estimate so the output approximates a normal distribution on $\R^d$. We sample a vector $\boldsymbol{z} \sim \mathcal{S}^{d-1}$ and compute the reward as:
\begin{equation*}
    F(s', \boldsymbol{z}) = \phi(s') \cdot \boldsymbol{z}.
\end{equation*}
Note that this is slightly different from the implementation in \Atari, where we divide by $\|\phi(s')\|$. We use reward normalization for RLE in both domains to scale the randomized reward, so both types have a similar effect.

\subsubsection{Evaluation Details in IsaacGym}\label{app:isaac_eval_details}

\paragraph{PPO normalized score} We use the IQM of the PPO normalized score to compute aggregate performance across 9 different environments in IsaacGym. We compute the PPO normalized score of the agent as $\mathrm{Agent}_{\mathrm{score}} / \mathrm{PPO}_{\mathrm{mean}}$. For example, the mean performance of PPO in a single environment under the PPO normalized score will be $1$. We compute the IQM of this metric for 5 seeds across 9 games (or 45 total runs) to aggregate performance.

\paragraph{Other evaluation details} Similar to our experiments in the \Atari domain, we use the bootstrapping method to estimate confidence intervals and use the probability of improvement to measure relative performance between different algorithms.

%% file: appx_further_visualizations.tex
\section{Visualizations on \textsc{FourRoom}}  \label{app:fourroom_visualizations}

In this section, we provide further results and visualizations for the \FourRoom environment: 

\begin{figure*}[t!]
\begin{center}    \includegraphics[width=\textwidth, clip]{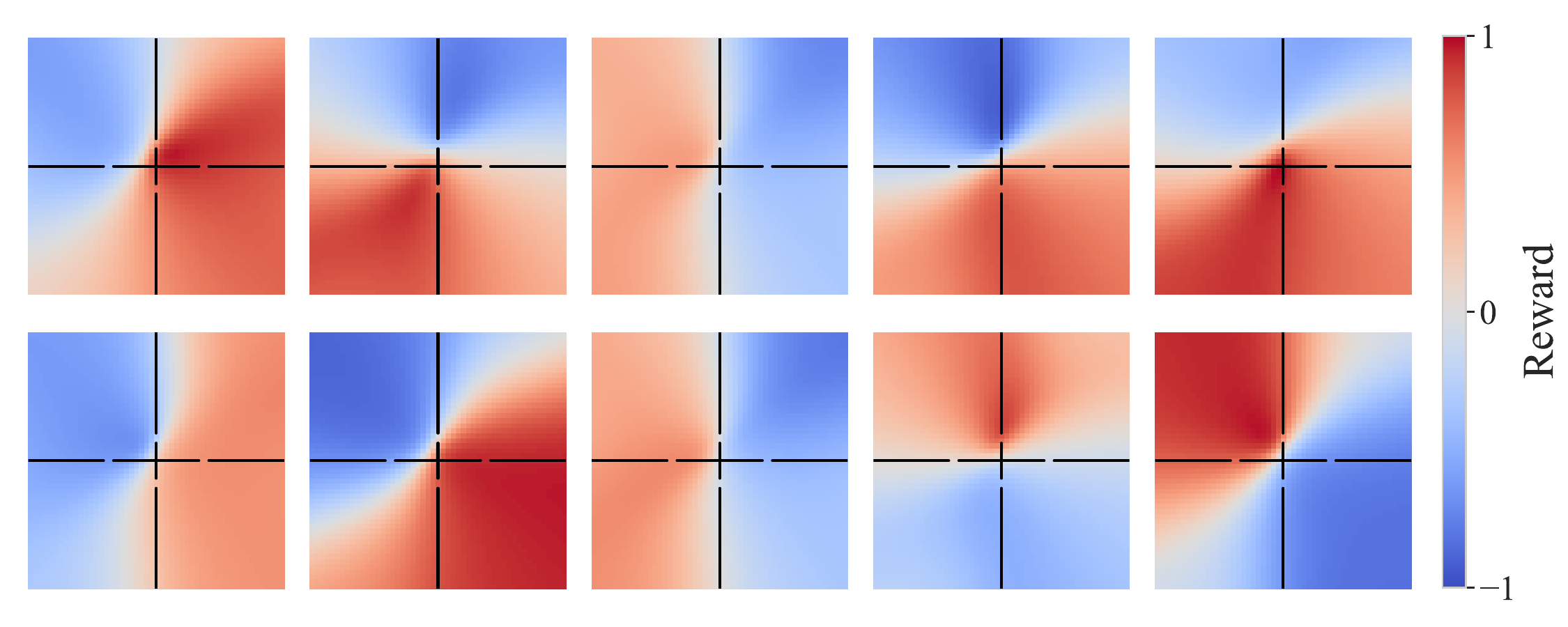}         
    \caption{Visualization of the reward function \(F(s; \boldsymbol{z})\) for 10 different random choices of \(\boldsymbol{z}\) in \FourRoom environment. The reward is is given by \(F(s, \boldsymbol{z}) = \boldsymbol{z} \cdot \phi(s) / \|\phi(s)\|\). The above image demonstrates the diversity and coverage of random reward functions in the \FourRoom environment.} 
    \label{app:fig:random_reward1}     
\end{center}
\end{figure*}  

\begin{figure*}[t!]
    \centering
    \begin{subfigure}{0.49\columnwidth}
        \centering
    \includegraphics[width=\columnwidth]{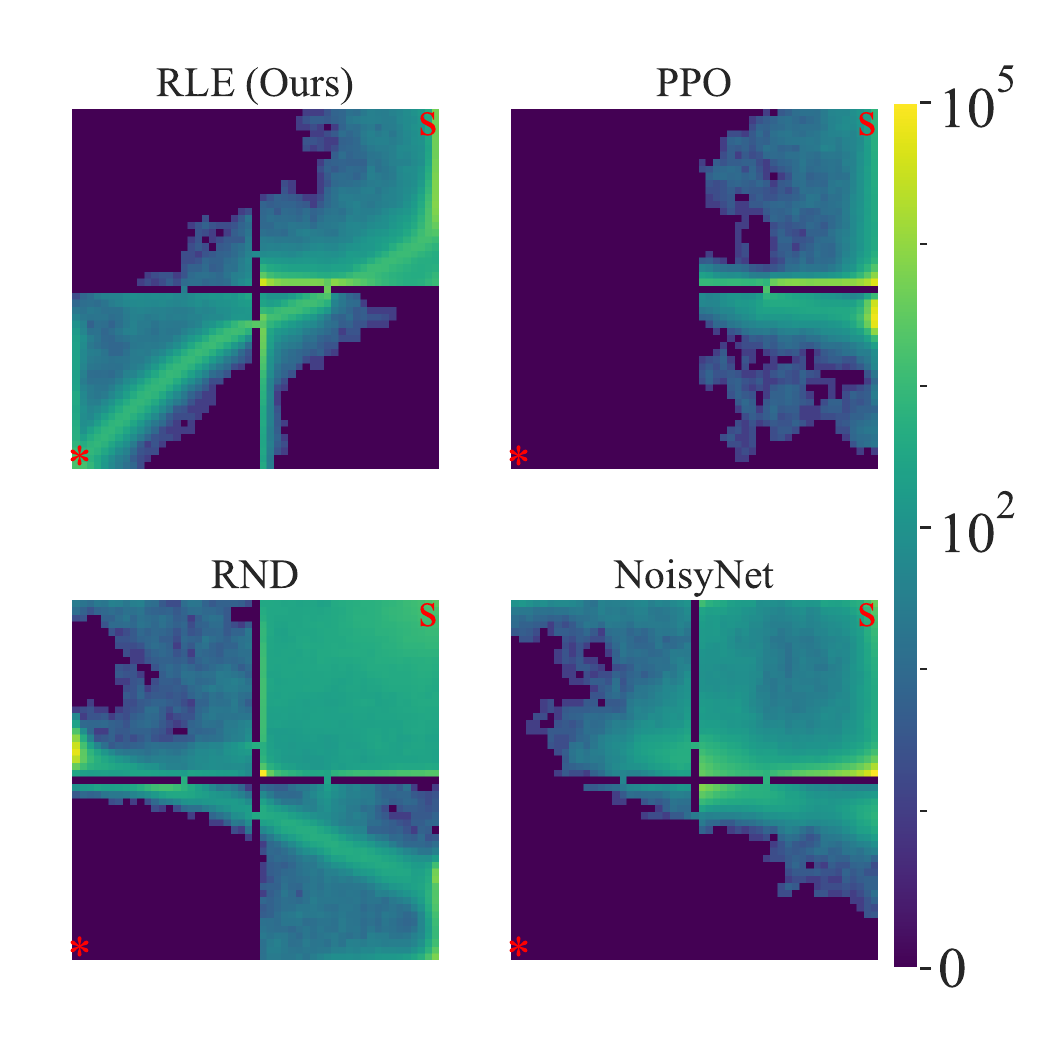}
    \caption{After 500K timesteps.}
    \end{subfigure}
    \begin{subfigure}{0.49\columnwidth}
        \centering    
    \includegraphics[width=\columnwidth]{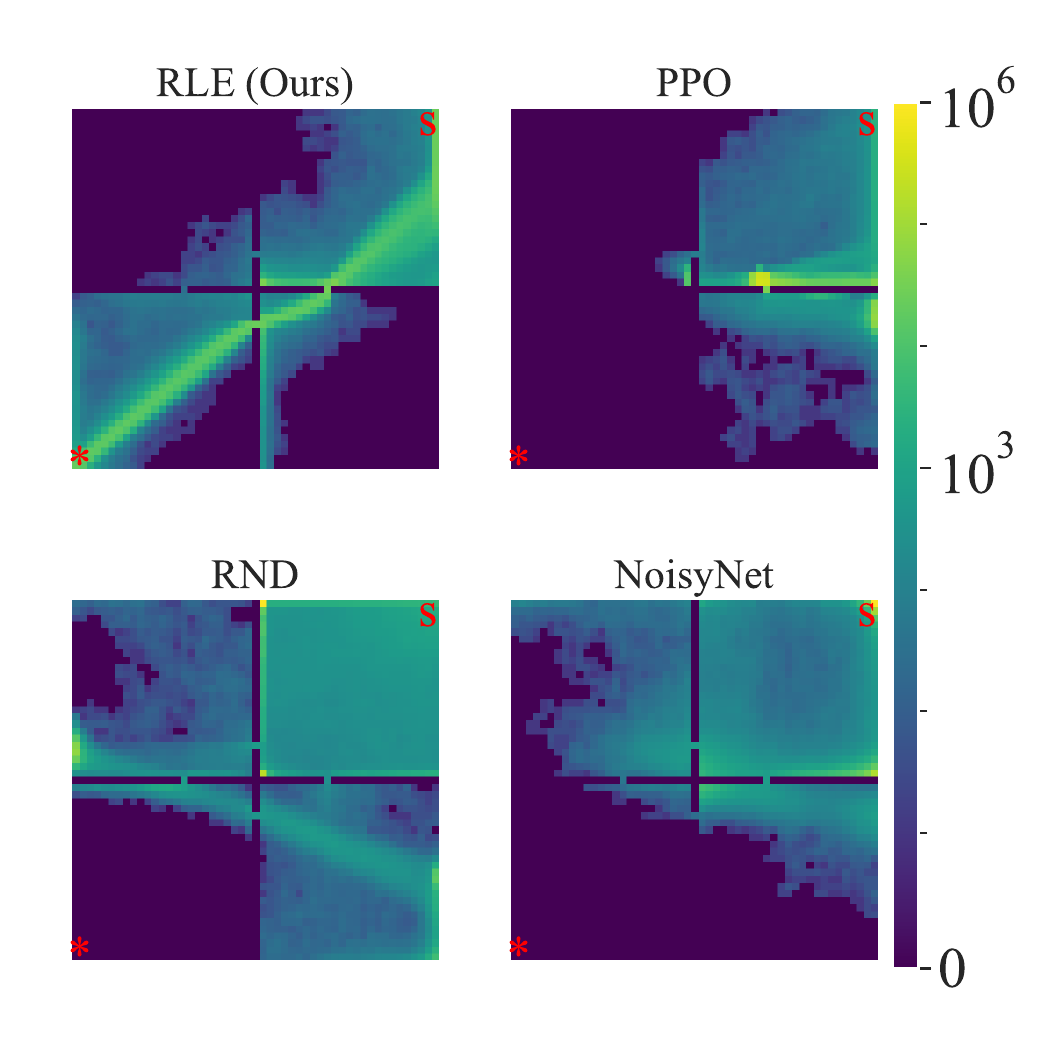}
        \caption{After 2.5M timesteps.}
    \end{subfigure}
    \caption{State visitation counts for different methods on \textsc{FourRooms} environment trained for 500K and 2.5M timesteps with task reward. The start location is the top-right state of the grid (represented by the red `S'). The agent gets a task reward of \(1\) at the bottom-left state (represented by red `*').} 
    \label{fig:app_visitation_ext} 
\end{figure*} 

\begin{figure*}[t!]
    \centering
    \begin{subfigure}{0.49\columnwidth}
        \centering
    \includegraphics[width=\textwidth]{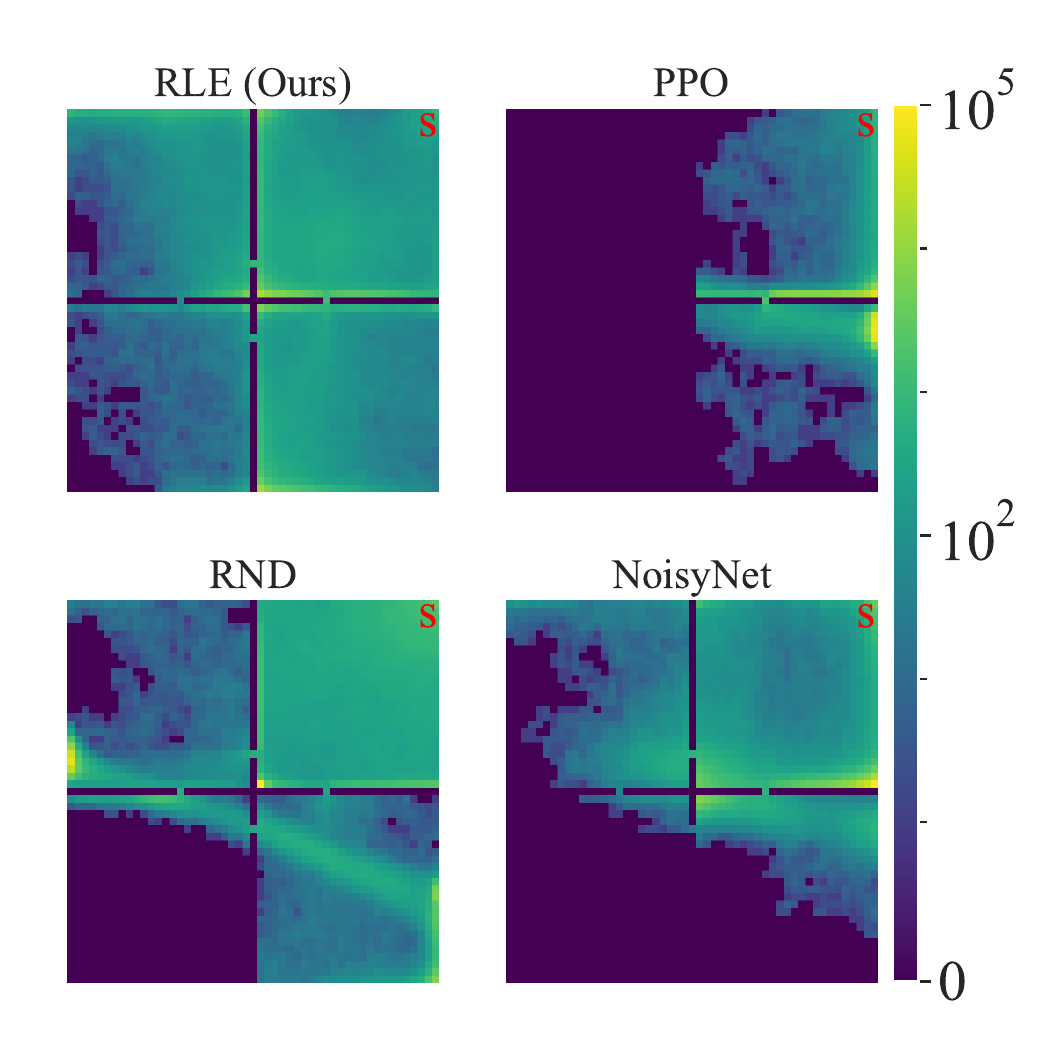} 
    \caption{After 500K steps}  
    \end{subfigure}%
    \begin{subfigure}{0.49\columnwidth}
        \centering
    \includegraphics[width=\textwidth]{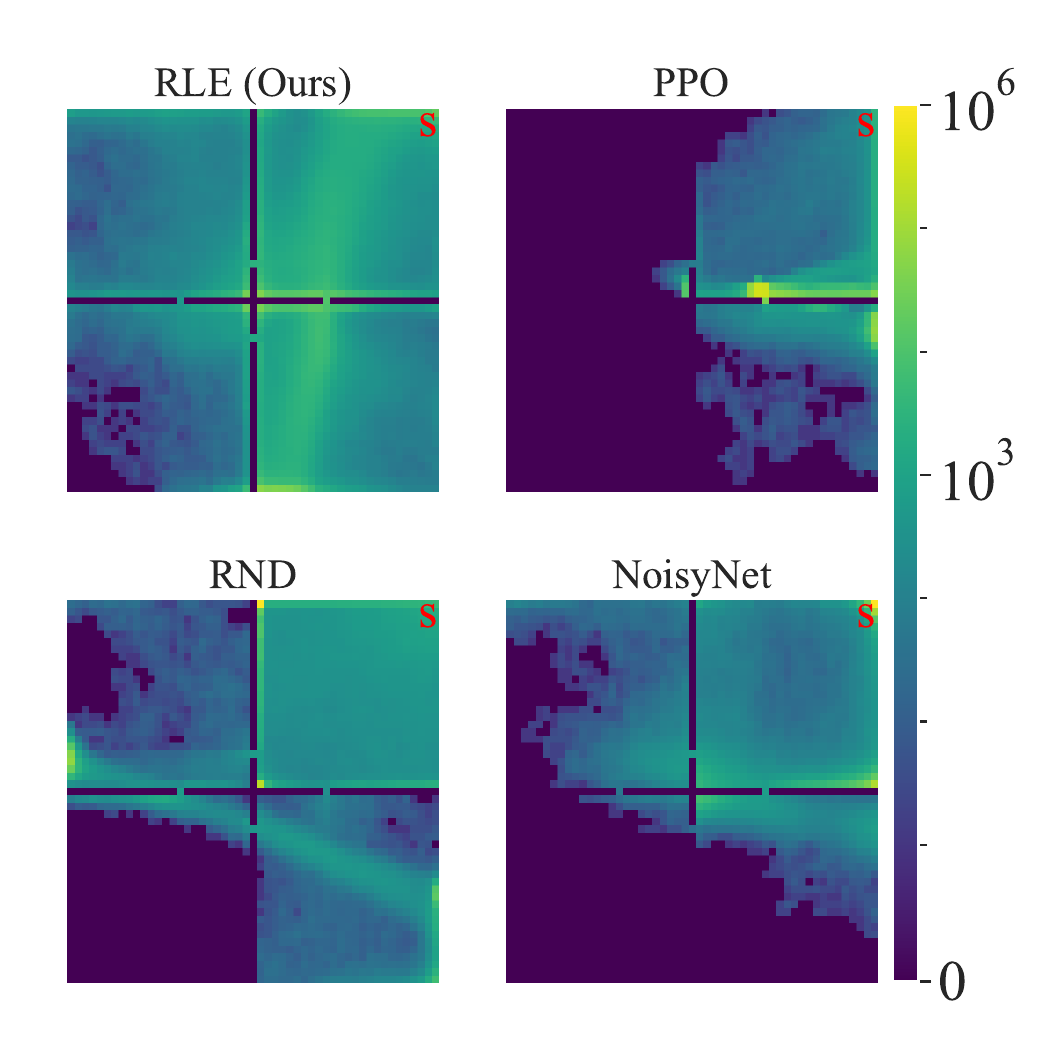}
        \caption{After 2.5M steps} 
    \end{subfigure}
    \caption{State visitation counts for different algorithms on \textsc{FourRooms} environment after training for 500K timesteps and 2.5M timesteps. All algorithms were trained without task reward (reward-free exploration).} 
    \label{fig:app_visitation_rf}
\end{figure*}

\begin{figure*}[t!]
    \centering
    \includegraphics[width=\textwidth]{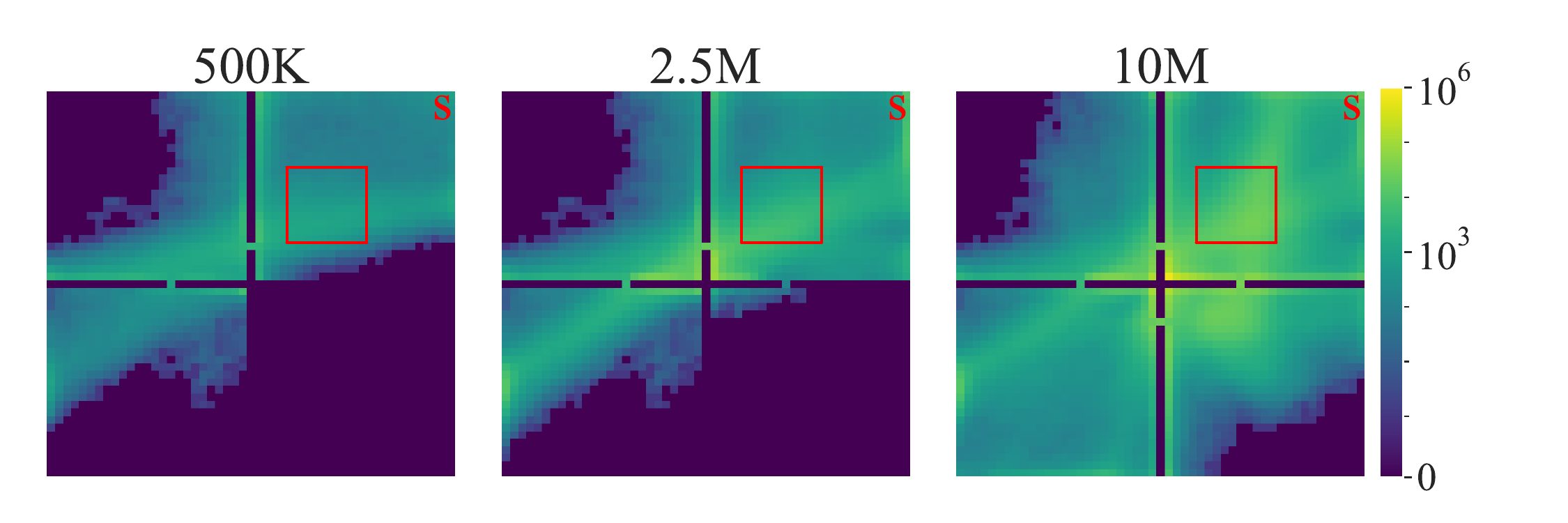}
    \caption{State visitation counts for RLE when trained in an environment with stochasticity in the observation space. The observation is only stochastic within the red square and is deterministic everywhere else. Even after discovering the red square, the agent is able to discover states outside of those regions and continues to explore throughout training. This suggests that RLE is less affected by the NoisyTV problem compared to novelty-based exploration methods.} 
    \label{fig:app_visitation_noisytv}
\end{figure*}

\paragraph{RL with task reward.}  In addition to the reward-free setting, we train all methods in the \FourRoom environment in a sparse-reward setting for 2.5M timesteps. There is a reward of $1$ in the bottom-left corner, and the reward is $0$ at all other states. We plot the state visitation counts of all methods after 500K and 2.5M timesteps in \Cref{fig:app_visitation_ext}. In addition, we train five seeds in this environment for each method, and find that the average score for \RLE and \NoisyNet is $0.6$, while the average score for \RND and \PPO is $0$. This suggests that the \FourRoom environment is a task that requires exploration as it is difficult for methods that rely on action noise like \PPO to achieve any reward.
\paragraph{State visitations.} 
\begin{itemize} 
\item Figure \ref{fig:app_visitation_ext} shows state visitation counts for all algorithms trained with a sparse task reward which is $1$ at the bottom-left state (red '*') and $0$ everywhere else.  
\item Figure \ref{fig:app_visitation_rf} shows state visitation counts for all algorithms trained for 500K and  2.5M steps without any task reward. 
\item \Cref{fig:app_visitation_noisytv} shows state visitation counts for RLE trained in a modified version of the environment with stochastic observations within a 2x2 square region of the environment. Through this, we test if RLE is susceptible to the ``NoisyTV'' problem~\cite{burda2018exploration}.
\end{itemize}

\begin{figure}
\begin{center}    
\includegraphics[width=0.65\textwidth]{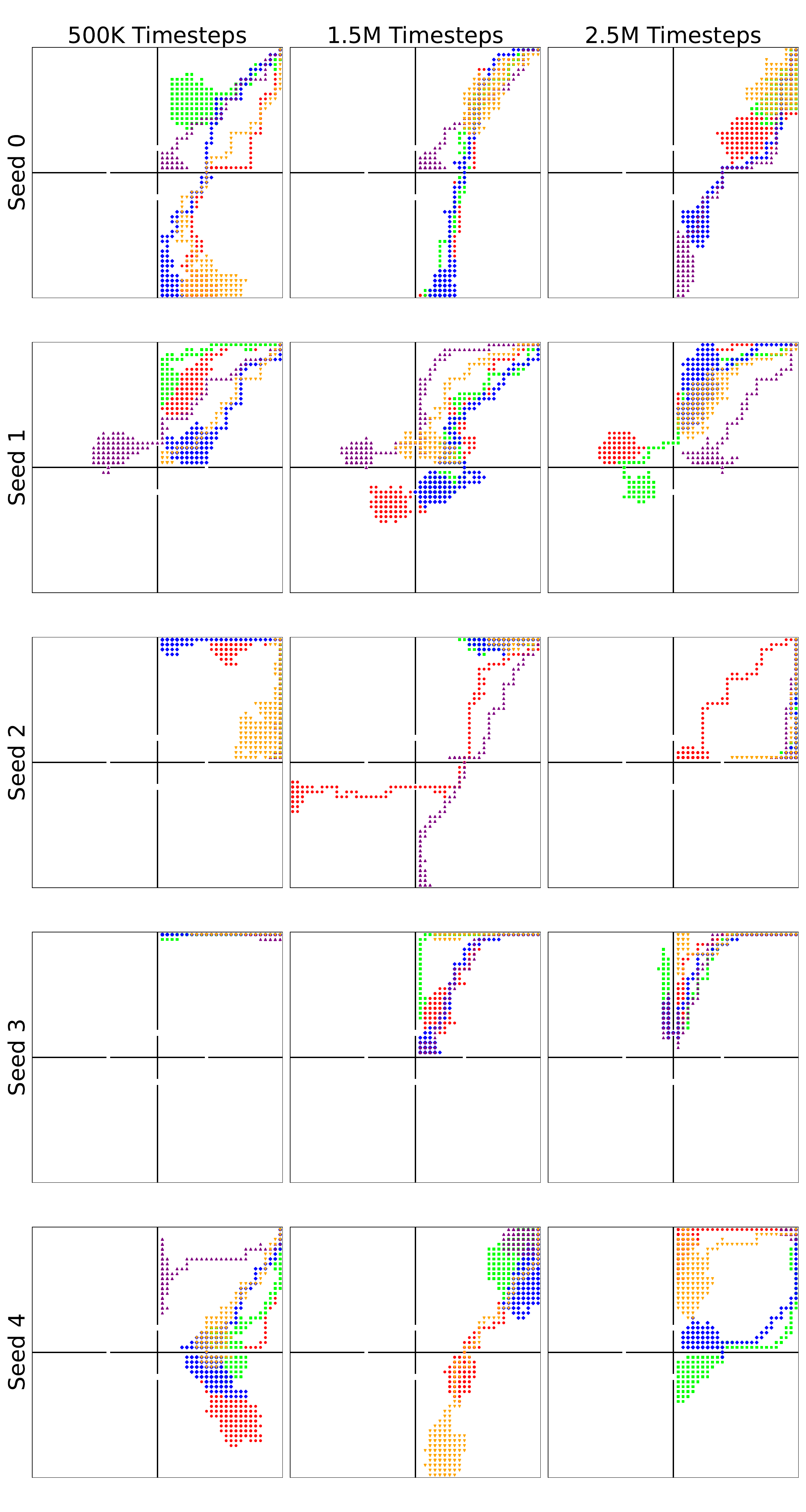}
\caption{Visualization of trajectories generated by sampling from a policy trained with \RLE for 2.5M timesteps in a reward-free setting across 5 seeds at different points in training. We sample 5 trajectories for each seed.} 
\label{app:fig:full_rle_trajectory_plots} 
\end{center} 
\end{figure}

\begin{figure}
\begin{center}    
\includegraphics[width=0.65\textwidth]{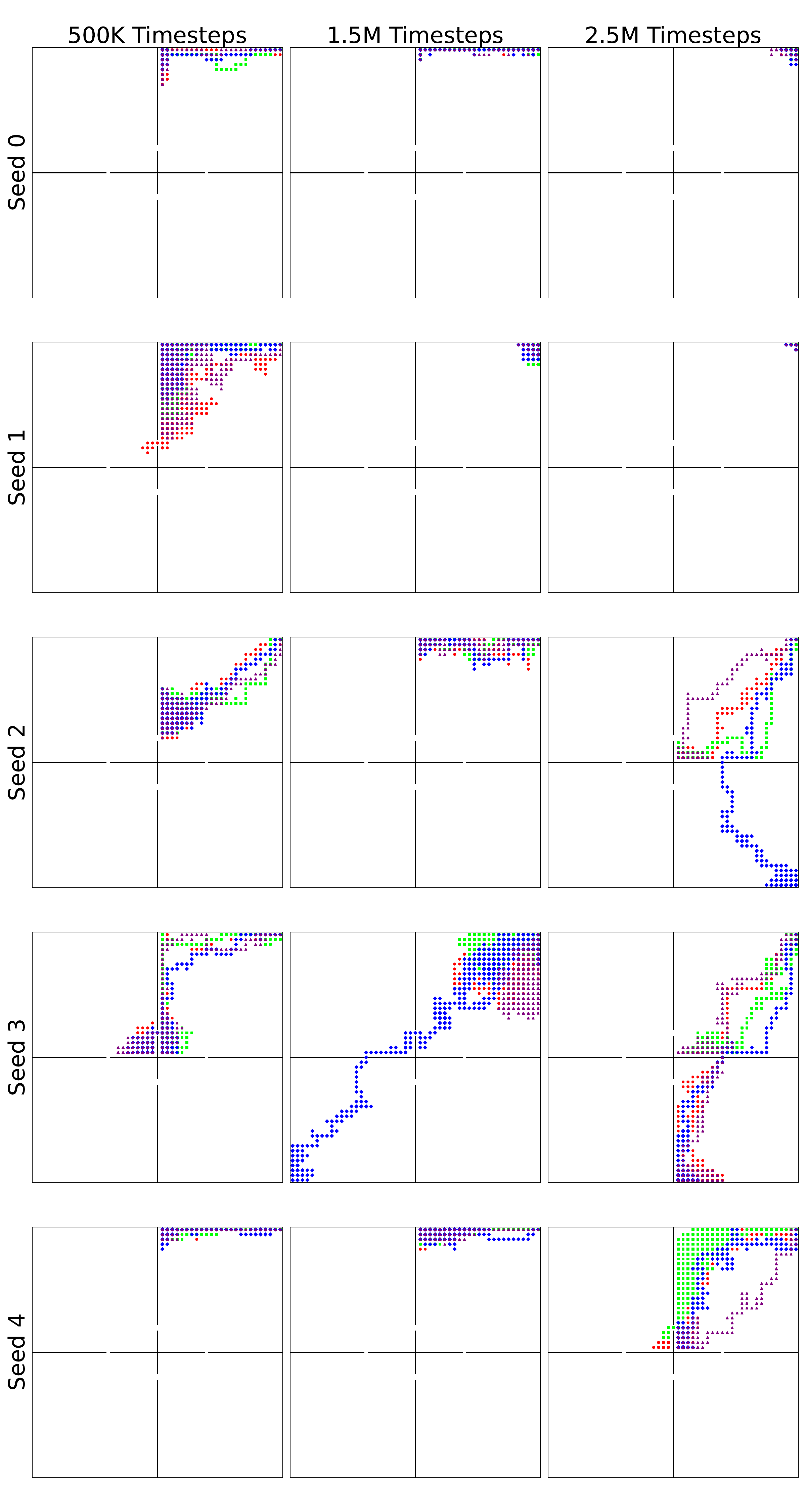}         
\caption{Visualization of trajectories generated by sampling from a policy trained with \NoisyNet for 2.5M timesteps in a reward-free setting across 5 seeds at different points in training. We sample 5 trajectories for each seed.}
\label{app:fig:full_noisy_net_trajectory_plots} 
\end{center} 
\end{figure}

\begin{figure}
\begin{center}    
\includegraphics[width=0.65\textwidth]{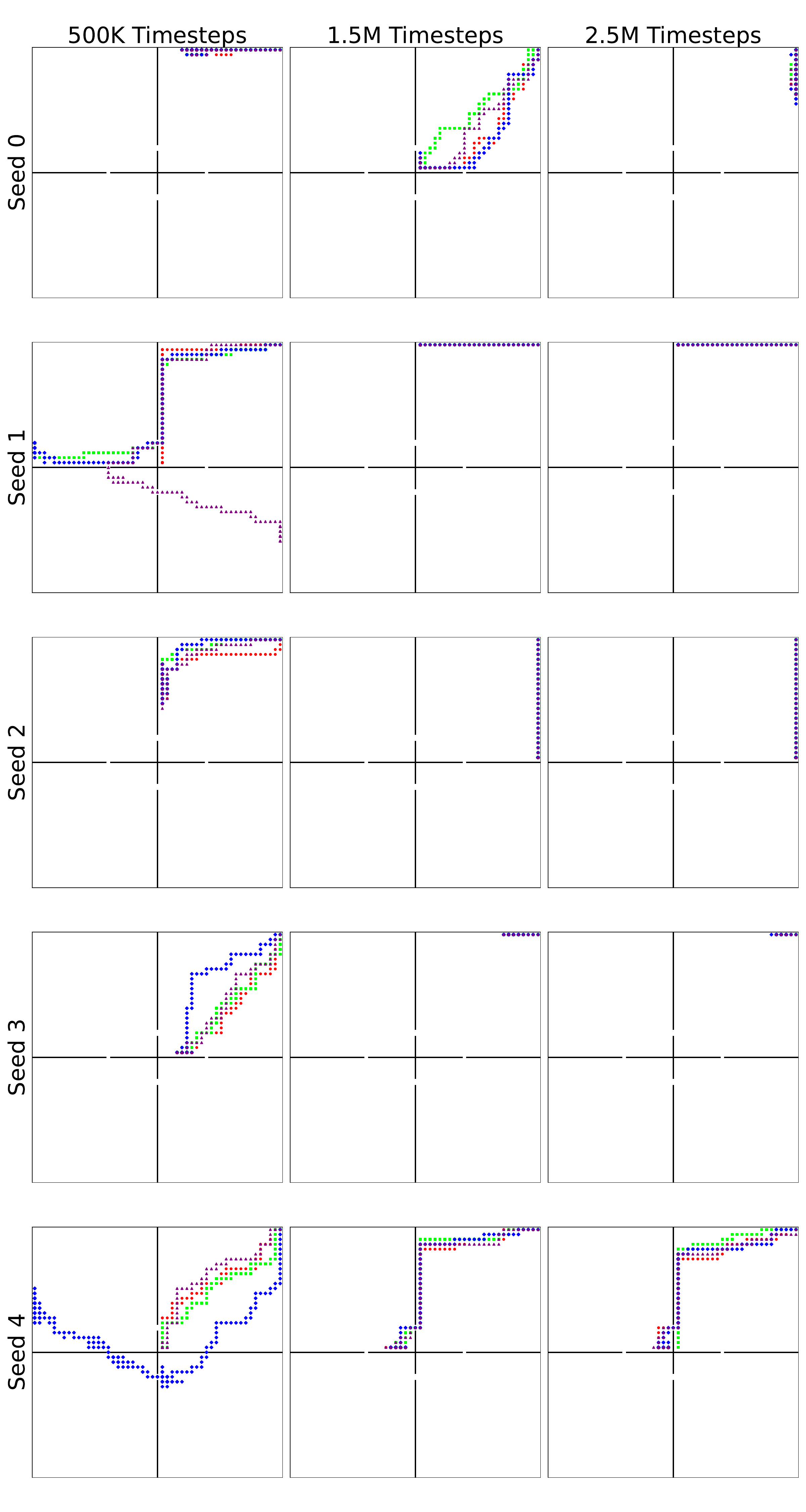}         
\caption{Visualization of trajectories generated by sampling from a policy trained with \RND for 2.5M timesteps in a reward-free setting across 5 seeds at different points in training. We sample 5 trajectories for each seed.}
\label{app:fig:full_rnd_trajectory_plots} 
\end{center} 
\end{figure}

\begin{figure}
\begin{center}    
\includegraphics[width=0.65\textwidth]{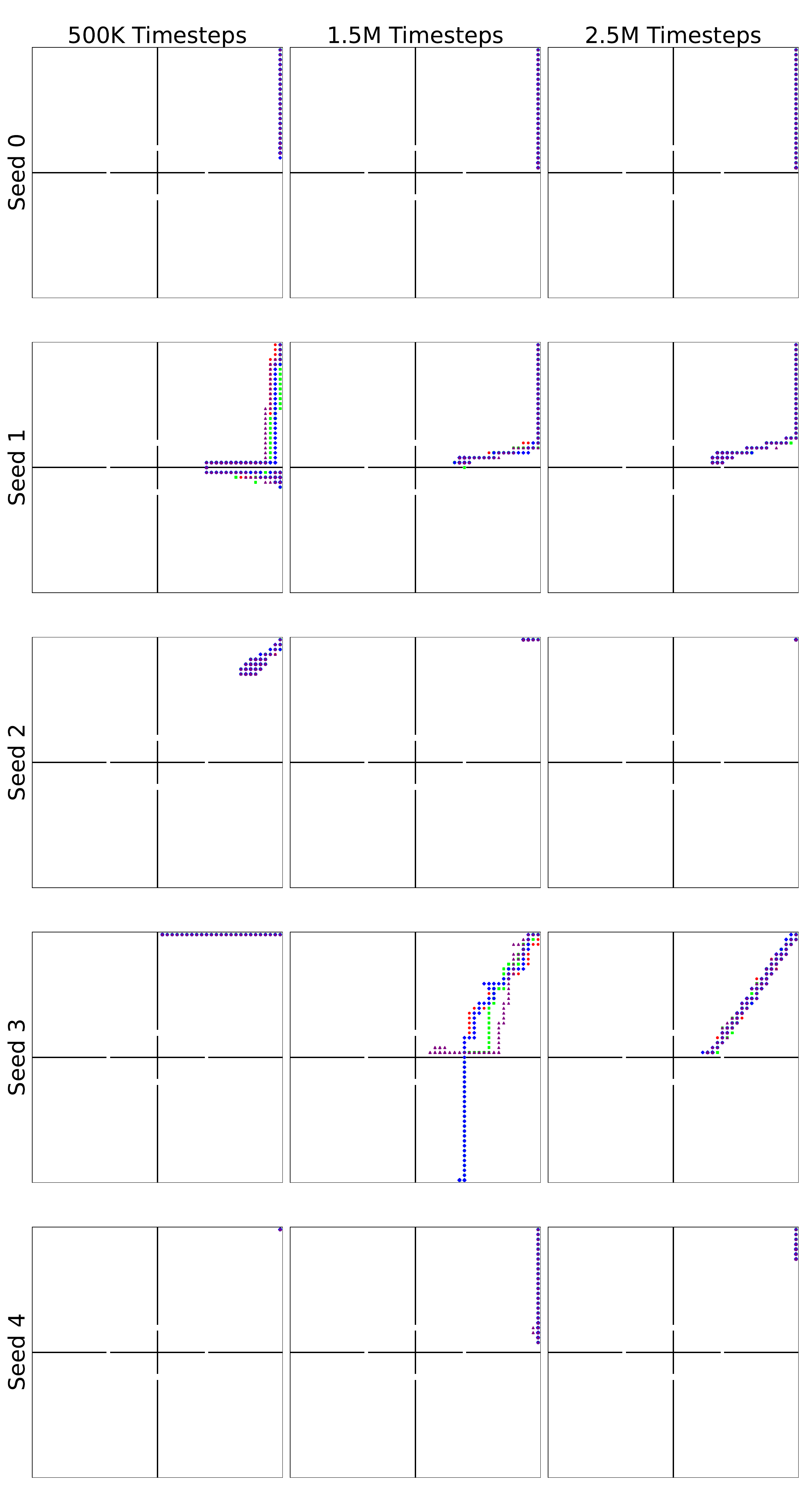}         
\caption{Visualization of trajectories generated by sampling from a policy trained with \PPO for 2.5M timesteps in a reward-free setting across 5 seeds at different points in training. We sample 5 trajectories for each seed.}
\label{app:fig:full_ppo_trajectory_plots} 
\end{center} 
\end{figure}

\paragraph{Visualizations of trajectory diversity across algorithms.} 
\begin{itemize}
    \item \Cref{app:fig:full_rle_trajectory_plots} shows 5 trajectories sampled from policies trained with \RLE across 5 different seeds at three different points in training: after 500K steps, 1.5M steps, and 2.5M steps.
    \item \Cref{app:fig:full_noisy_net_trajectory_plots} shows the same for \NoisyNet.
    \item \Cref{app:fig:full_rnd_trajectory_plots} shows the same for \RND.
    \item \Cref{app:fig:full_ppo_trajectory_plots} shows the same for \PPO.
\end{itemize}

From visual evaluation, the above plots suggest that \RLE induces more diverse trajectories as compared to other baselines (\PPO, \RND, and \NoisyNet) on the \FourRoom environment. 
 
%% file: detailed_results.tex
\clearpage 

\section{Detailed Results and Learning Curves on all \textsc{Atari} Games} 
\label{app:sec:atari_detailed}
\begin{figure}
\begin{center}    
\includegraphics[width=0.95\textwidth] 
{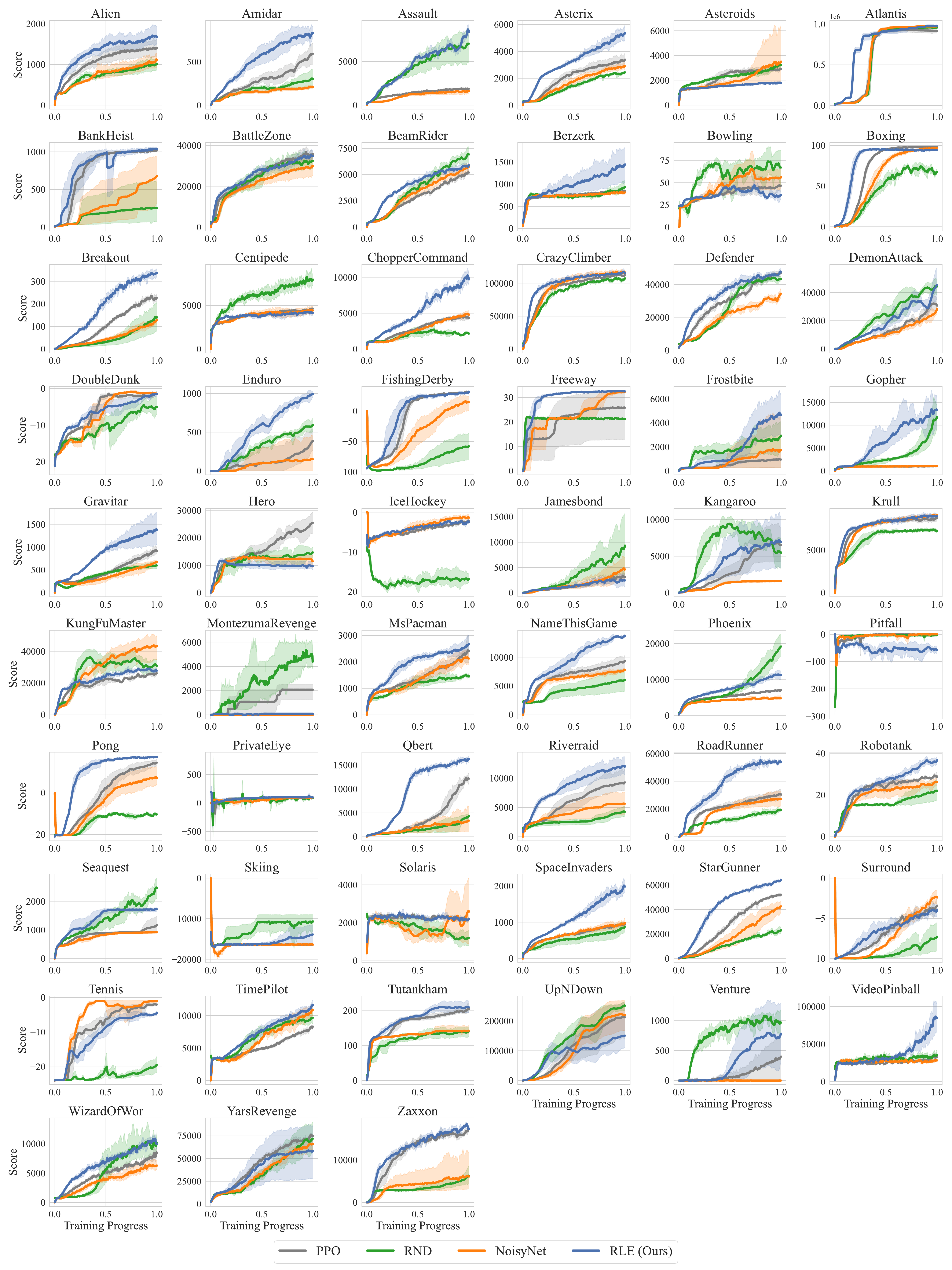}
\caption{Game scores for different algorithms for all 57 \Atari games.}     \label{fig:atari_learning_curves} 
\end{center} 
\end{figure}

\begin{figure}
\begin{center}    
\includegraphics[width=0.6\textwidth]{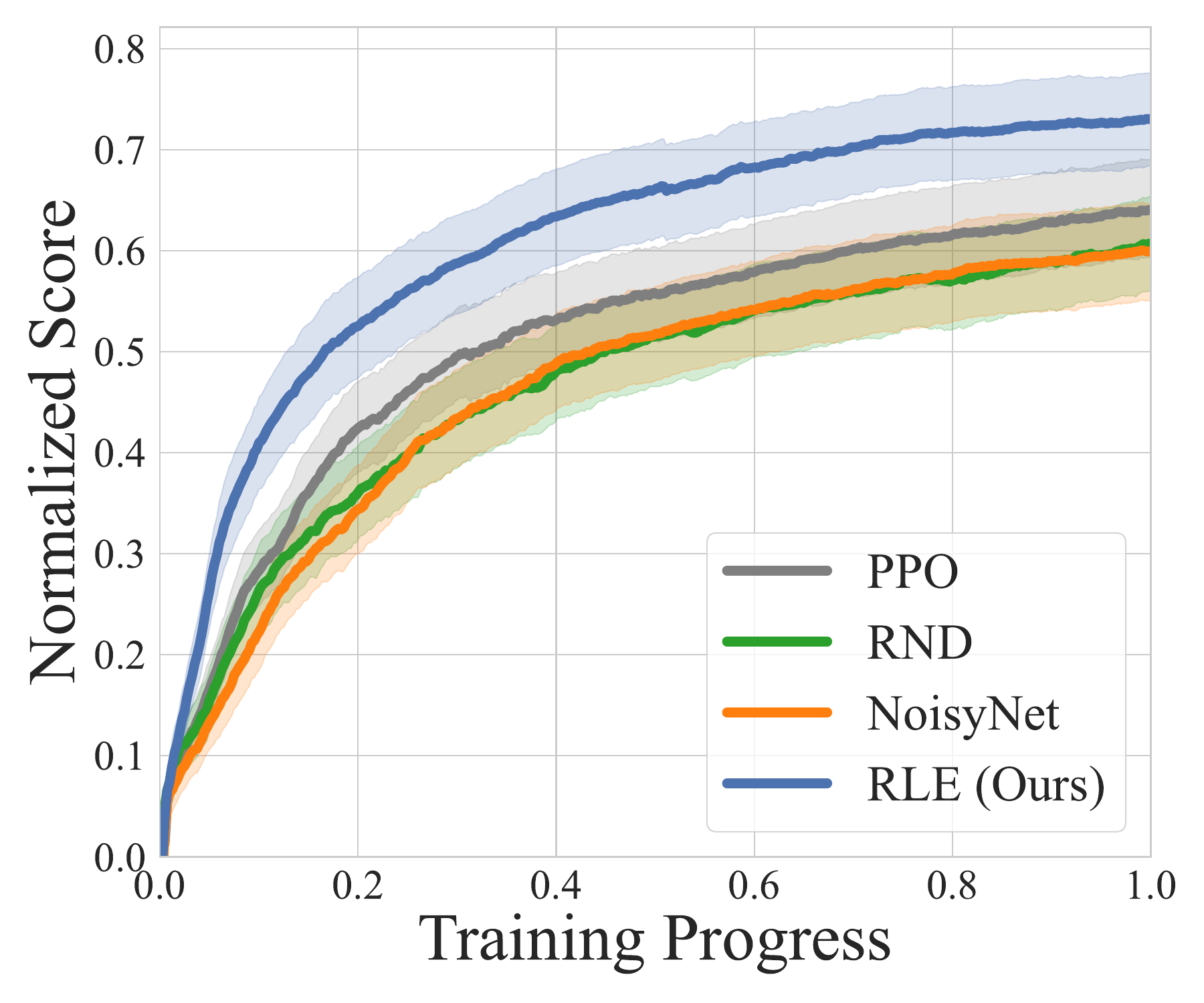}
\caption{Capped human normalized score across all 57 Atari games. RLE outperforms all other methods in this metric and requires half the training time to reach the same score as the next best method (PPO).}     \label{fig:atari_chns_mean}  
\end{center} 
\end{figure}

\begin{figure}
    \begin{center}
        \begin{subfigure}{0.45\textwidth}
            \centering
            \includegraphics[width=\textwidth]{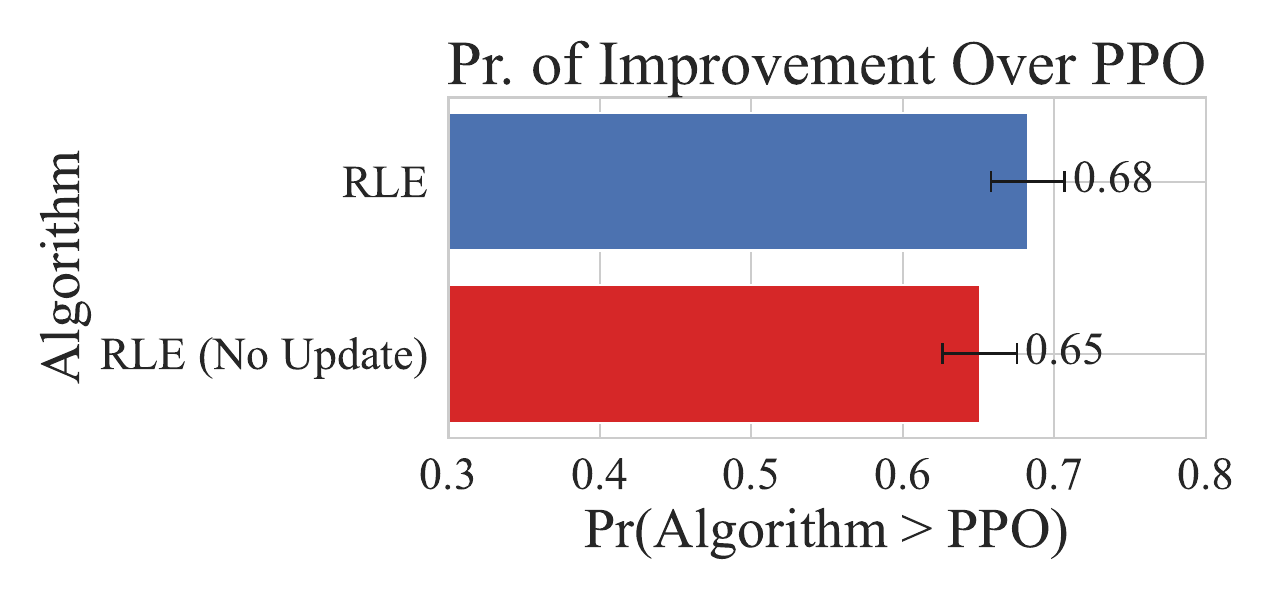}
            \caption{Probability of improvement over PPO with and without a slow value feature update rule. Using the value features leads to a slight increase in performance.}
            \label{fig:poi_ablation_over_ppo}
        \end{subfigure}
        \hfill
        \begin{subfigure}{0.45\textwidth}
            \centering
            \includegraphics[width=\textwidth]{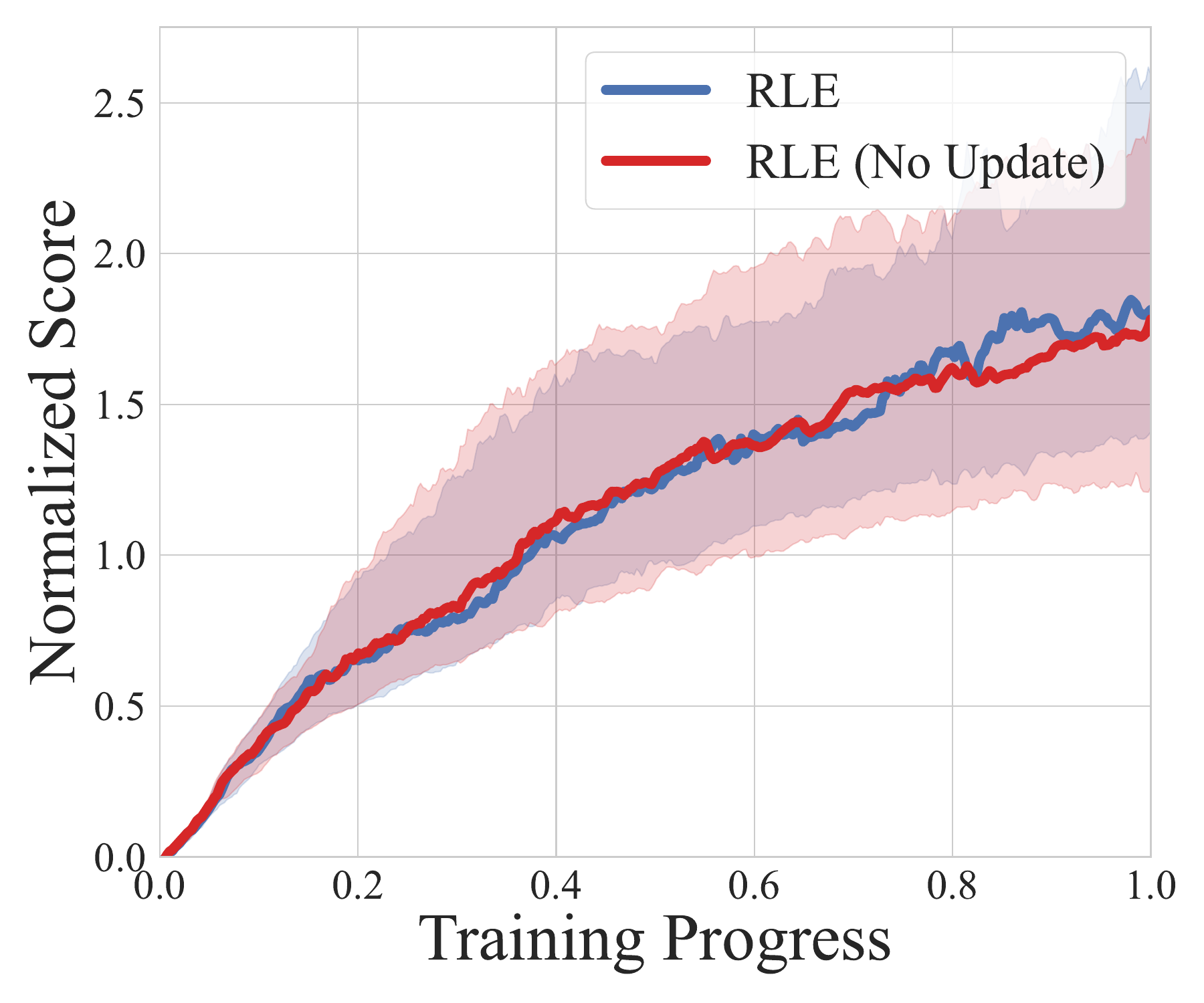}
            \caption{IQM of human normalized score of RLE, both with and without a slow value feature update rule. With respect to this metric, both versions of the method perform very similarly overall.}
            \label{fig:aggregated_hns_iqm}
        \end{subfigure}
        \caption{Comparison of RLE performance with and without a slow value feature update rule.}
        \label{fig:atari_value_feat_combined_ablation}
    \end{center}
\end{figure}

\begin{figure}
\begin{center}    
\includegraphics[width=0.95\textwidth]{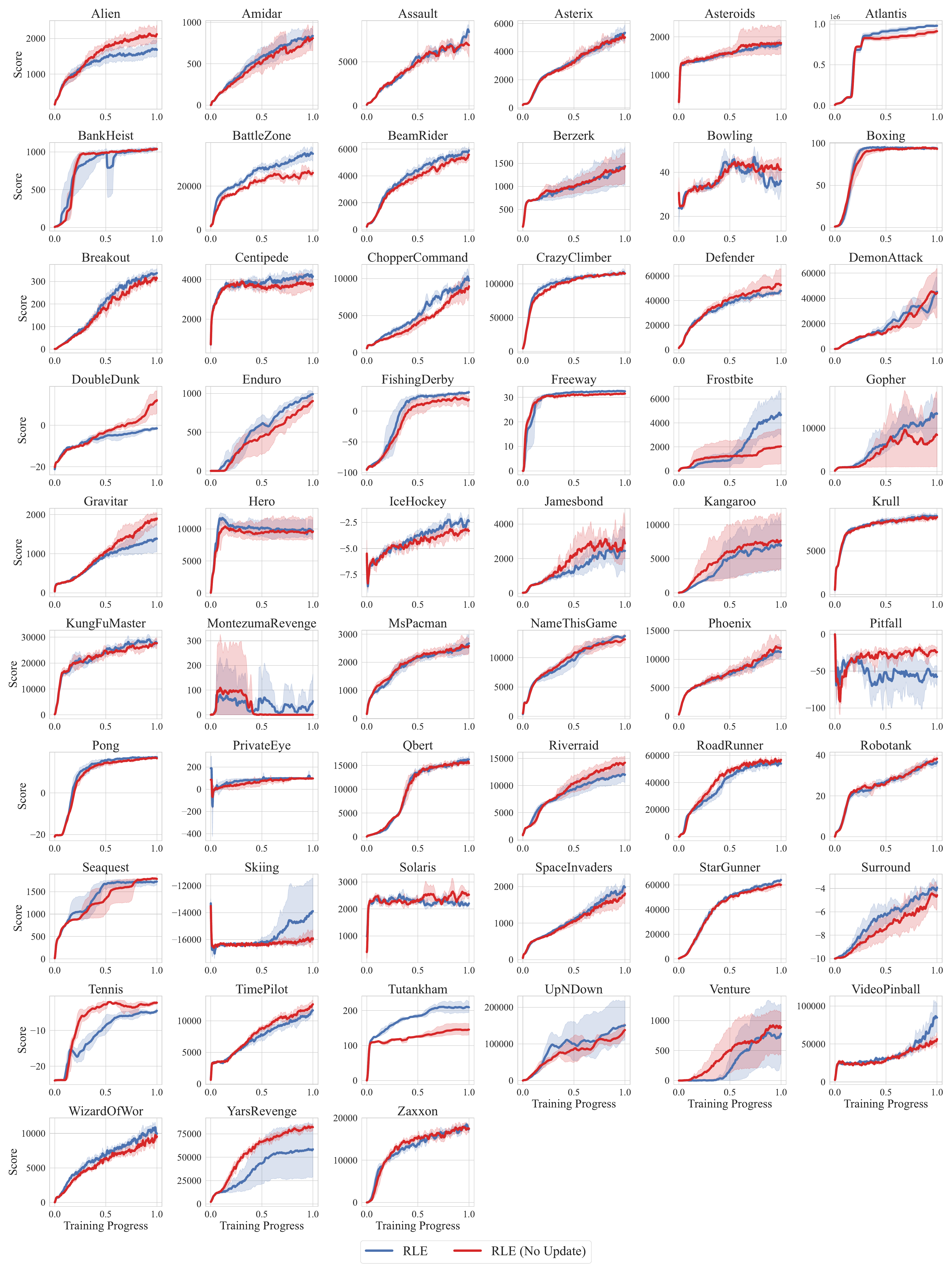}
\caption{Game scores for our method with and without a slow value feature update rule. Performance is usually similar, but noticeably different in a handful of games (for example Alien, Frostbite, Skiing, Tutankham, YarsRevenge).}     \label{fig:atari_value_feat_all_games_ablation}
\end{center} 
\end{figure}

We provide: 
\begin{itemize}
    \item The scores for each of the algorithms ({\RLE (Ours), \PPO, \RND and \NoisyNet}) on all 57 \Atari games in Table \ref{tab:atari_results}. Each of the algorithms was trained for 40M steps on all Atari games except for the results for \textsc{Montezuma's Revenge} where we trained for 400M steps. The reported Human performance is obtained from \citet{mnih2013playing, badia2020agent57}.
    \item Learning curves for all the algorithms ({\RLE (Ours), \PPO, \RND and \NoisyNet}) for all 57 \Atari games in Figure \ref{fig:atari_learning_curves}. 
    \item Aggregated capped human normalized score (described in \Cref{app:atari_eval_details}) for each of the algorithms ({\RLE (Ours), \PPO, \RND and \NoisyNet}) over all 57 Atari games in Figure \ref{fig:atari_chns_mean}
    \item An ablation study of how the soft update rule for the feature network affects performance on \Atari games taking three metrics into account (scores on individual games, IQM of human normalized score, and probability of improvement over PPO). In the \Atari experiments described in Section~\ref{subsec:exp:result}, we used a slow-moving estimate of the CNN features  (see \Cref{subsubsec:rle_imp_details_atari}) learned by the value network to compute \RLE rewards. This choice of features slightly contributes to improved performance. \Cref{fig:atari_value_feat_combined_ablation} presents the IQM of the normalized score and the POI of \RLE over PPO using the CNN features learned by the value network vs. using the randomly initialized CNN features. The results demonstrate that incorporating value network features leads to a higher POI, while not significantly affecting the IQM. We plot the learning curve in each game for both variants in \Cref{fig:atari_value_feat_all_games_ablation}, finding that while performance is broadly similar, there are a few games where the two variants have significant differences in performance.
\end{itemize} 

\renewcommand{\arraystretch}{0.85}
\begin{table}
\begin{center}
\begin{tabular}{lrrrr}
\toprule
{} &         PPO &         RND &    NoisyNet &  RLE (Ours) \\
\midrule
Alien-v5 & 1409.22 & 1010.94 & 1112.59 & \textbf{1680.13} \\
Amidar-v5 & 595.53 & 304.01 & 210.88 & \textbf{836.01} \\
Assault-v5 & 1886.68 & 7045.97 & 1605.63 & \textbf{8368.57} \\
Asterix-v5 & 3392.84 & 2439.56 & 2892.50 & \textbf{5350.79} \\
Asteroids-v5 & 2913.15 & 3232.00 & \textbf{3513.02} & 1798.64 \\
Atlantis-v5 & 915703.86 & 957537.41 & 972286.70 & \textbf{979023.58} \\
BankHeist-v5 & 1017.55 & 250.91 & 678.44 & \textbf{1036.61} \\
BattleZone-v5 & \textbf{35832.79} & 32536.65 & 30021.64 & 34793.65 \\
BeamRider-v5 & 5200.10 & \textbf{6915.42} & 5755.02 & 5895.93 \\
Berzerk-v5 & 845.95 & 931.95 & 812.45 & \textbf{1445.49} \\
Bowling-v5 & 46.84 & \textbf{67.17} & 56.17 & 35.88 \\
Boxing-v5 & \textbf{97.99} & 67.44 & 97.06 & 93.94 \\
Breakout-v5 & 227.34 & 139.41 & 127.66 & \textbf{337.07} \\
Centipede-v5 & 4634.04 & \textbf{7972.15} & 4419.47 & 4151.10 \\
ChopperCommand-v5 & 4342.80 & 2140.08 & 4787.63 & \textbf{9710.00} \\
CrazyClimber-v5 & 112278.70 & 107602.05 & \textbf{116637.17} & 115593.01 \\
Defender-v5 & 46957.65 & 43387.18 & 34448.07 & \textbf{47872.91} \\
DemonAttack-v5 & 30714.39 & 44342.27 & 28287.09 & \textbf{45217.82} \\
DoubleDunk-v5 & -1.57 & -5.04 & \textbf{-1.43} & -1.51 \\
Enduro-v5 & 387.89 & 595.65 & 150.38 & \textbf{990.95} \\
FishingDerby-v5 & \textbf{31.13} & -57.88 & 14.14 & 30.68 \\
Freeway-v5 & 25.83 & 21.19 & 32.40 & \textbf{32.49} \\
Frostbite-v5 & 949.08 & 2944.37 & 1747.85 & \textbf{4658.90} \\
Gopher-v5 & 1020.40 & 11822.26 & 1055.82 & \textbf{13290.12} \\
Gravitar-v5 & 920.19 & 597.42 & 674.99 & \textbf{1381.69} \\
Hero-v5 & \textbf{25495.80} & 14695.30 & 11433.06 & 9668.68 \\
IceHockey-v5 & -2.09 & -16.70 & \textbf{-1.34} & -2.39 \\
Jamesbond-v5 & 3157.81 & \textbf{9347.30} & 4633.37 & 2452.21 \\
Kangaroo-v5 & 6504.67 & 5474.45 & 1596.99 & \textbf{6992.13} \\
Krull-v5 & 8731.23 & 7264.60 & \textbf{9063.52} & 8981.43 \\
KungFuMaster-v5 & 26131.84 & 30902.44 & \textbf{43341.34} & 27813.32 \\
MontezumaRevenge-v5 & 2077.03 & \textbf{4406.79} & 0.00 & 79.48 \\
MsPacman-v5 & 2417.82 & 1446.32 & 2127.62 & \textbf{2676.20} \\
NameThisGame-v5 & 9392.45 & 6078.34 & 7818.62 & \textbf{13701.36} \\
Phoenix-v5 & 7137.14 & \textbf{19195.54} & 4786.92 & 11272.80 \\
Pitfall-v5 & -0.67 & -3.41 & \textbf{-0.05} & -57.65 \\
Pong-v5 & 14.52 & -10.34 & 7.10 & \textbf{17.17} \\
PrivateEye-v5 & \textbf{98.34} & 87.24 & 95.55 & 97.79 \\
Qbert-v5 & 12168.36 & 4300.73 & 3381.40 & \textbf{16261.59} \\
Riverraid-v5 & 9268.85 & 4267.51 & 5642.73 & \textbf{12009.63} \\
RoadRunner-v5 & 30354.36 & 19452.68 & 27037.68 & \textbf{53920.12} \\
Robotank-v5 & 28.68 & 22.11 & 26.34 & \textbf{36.71} \\
Seaquest-v5 & 1172.22 & \textbf{2463.42} & 920.71 & 1724.96 \\
Skiing-v5 & -16370.14 & \textbf{-10644.07} & -16398.72 & -13887.77 \\
Solaris-v5 & 2203.41 & 1206.94 & \textbf{2584.66} & 2203.76 \\
SpaceInvaders-v5 & 938.00 & 878.91 & 981.24 & \textbf{1981.37} \\ 
StarGunner-v5 & 52219.39 & 23174.16 & 42645.43 & \textbf{64011.13} \\
Surround-v5 & -3.41 & -7.28 & \textbf{-2.41} & -3.91 \\
Tennis-v5 & -2.03 & -19.44 & \textbf{-1.06} & -4.49 \\
TimePilot-v5 & 8319.51 & 9695.60 & 10888.94 & \textbf{11636.24} \\
Tutankham-v5 & 204.57 & 140.97 & 142.70 & \textbf{209.23} \\
UpNDown-v5 & 212171.29 & \textbf{251442.66} & 219951.36 & 151036.48 \\
Venture-v5 & 401.20 & \textbf{969.00} & 0.06 & 782.98 \\
VideoPinball-v5 & 32654.24 & 35275.58 & 28236.75 & \textbf{84825.64} \\
WizardOfWor-v5 & 8355.05 & \textbf{10151.69} & 6306.86 & 9942.29 \\
YarsRevenge-v5 & \textbf{74833.17} & 71789.37 & 65902.61 & 58507.98 \\
Zaxxon-v5 & 17354.21 & 6273.86 & 6104.86 & \textbf{17403.15} \\
\bottomrule
\end{tabular}
\end{center}

\caption{Performance on all 57 \Atari games. Each algorithm was trained for 40M timesteps, except for \textsc{Montezuma's Revenge} where we trained for 400M timesteps. The reported Human performance is obtained from \citet{mnih2013playing, badia2020agent57}.}  
\label{tab:atari_results} 
\end{table}

\section{Detailed Results and Learning Curves on all \IsaacGym Tasks} 
\begin{figure*}[htb!]
    \centering
    \includegraphics[width=0.9\textwidth]{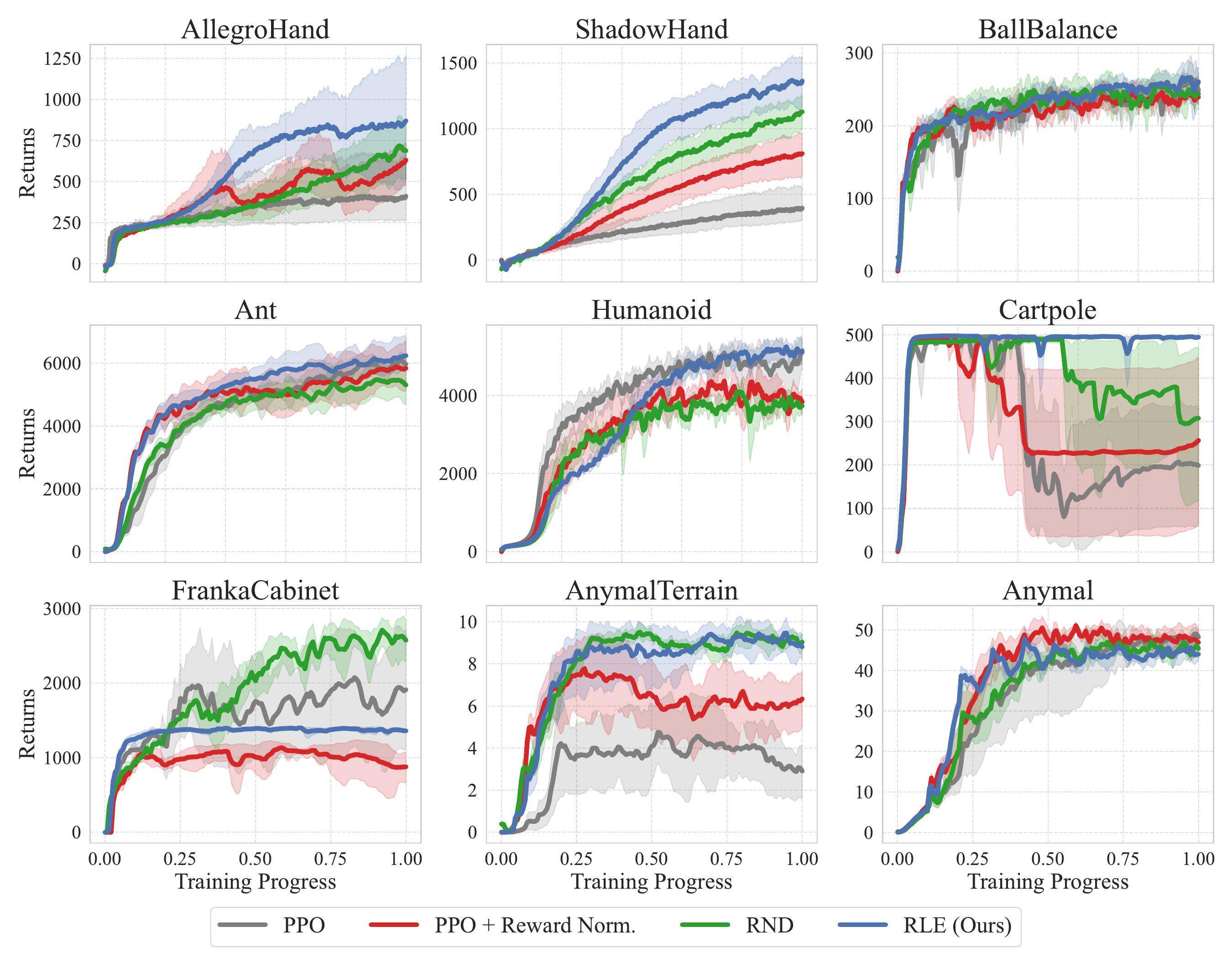}    \caption{Comparison of achieved returns between \RLE and standard \PPO (higher is better). \RLE achieves return greater than or equal to that of standard \PPO in the majority of tasks. We also compare \RLE to an ablation of \PPO that uses reward normalization and find that \RLE improves over it as well. Finally, we compare \RLE to \RND, finding that while \RND surprisingly improves performance in these tasks compared to \PPO, \RLE improves over \RND in the majority of tasks.} 
    \label{fig:result:all_isaacgym}
\end{figure*}

\begin{figure}[t!]
    \centering
    \includegraphics[width=0.9\textwidth]{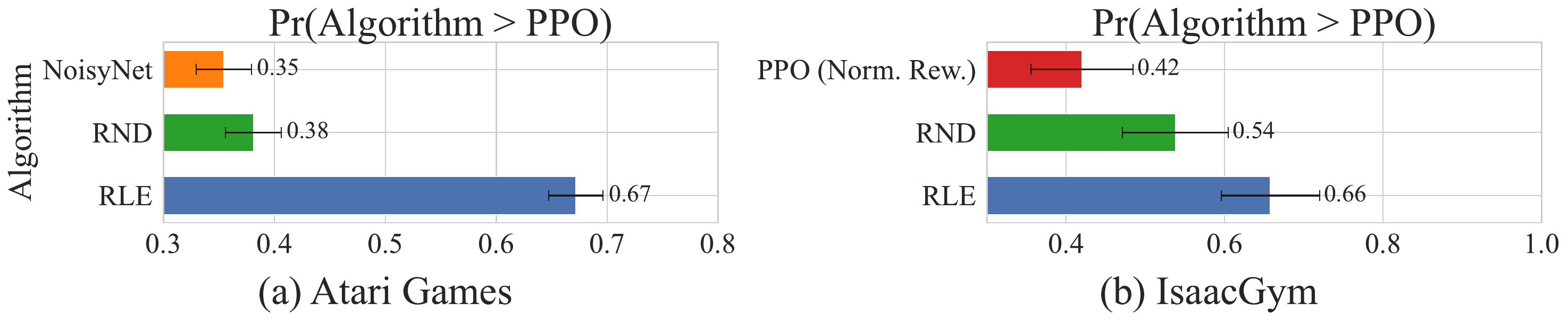}
    \caption{
    \textbf{(a)} Probability of improvement of \RLE, \RND, and \NoisyNet over \PPO across all $57$ \Atari games. POI over \PPO of both \NoisyNet and \RND are below $0.5$, implying that neither \NoisyNet nor \RND statistically improve over \PPO overall across $57$ \Atari games.
    \textbf{(b)} Probability of improvement of \RLE, \RND, and \PPO with reward normalization over \PPO across all $9$ \IsaacGym tasks. POI over \PPO of both \RND and \PPO with reward normalization are below $0.5$, demonstrating that neither baseline statistically improves over \PPO overall across the \IsaacGym tasks.}
    \label{fig:result:poi_over_ppo}
\end{figure}

\begin{figure}[htb!]
    \centering
    \includegraphics[width=0.5\textwidth]{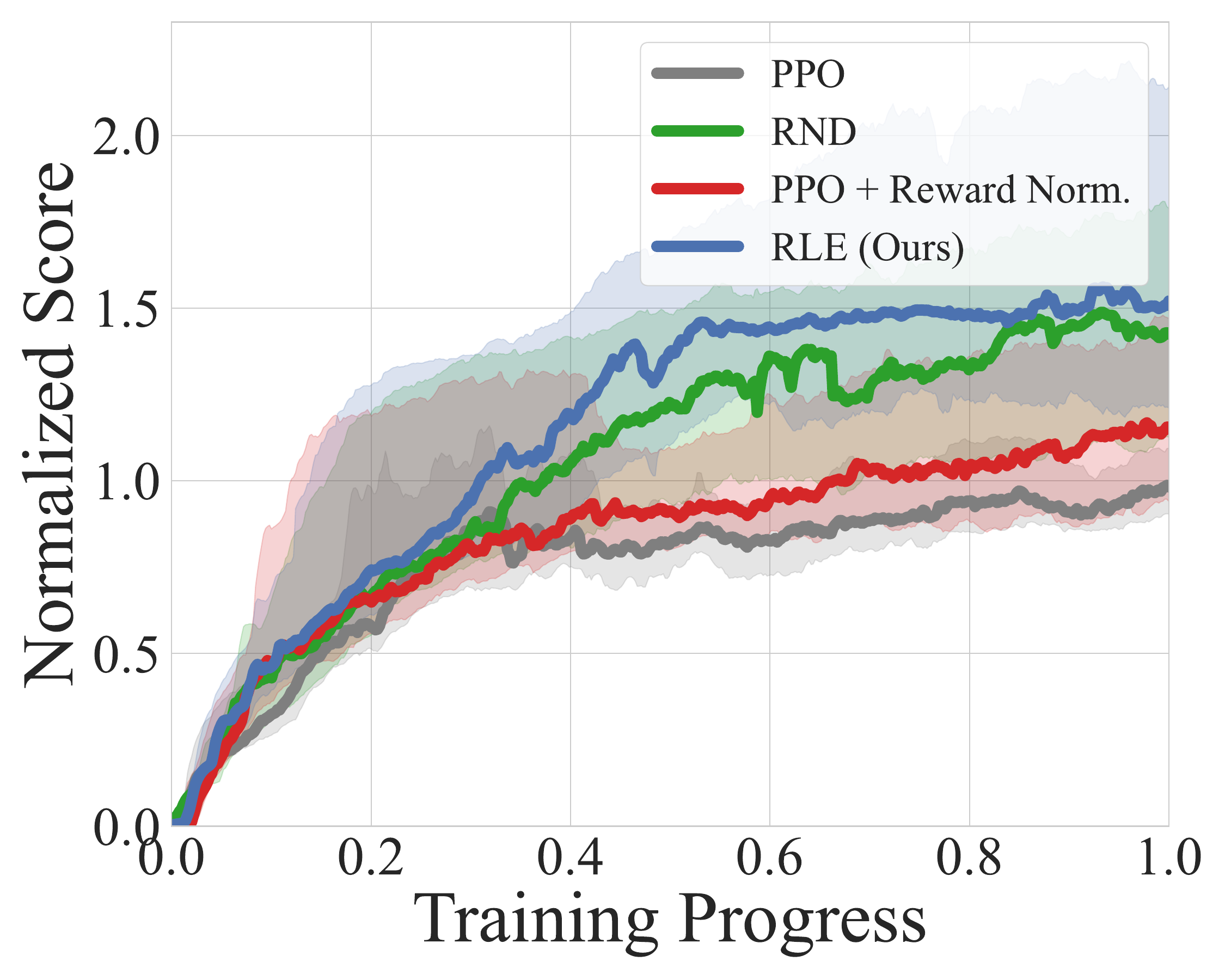}
    \caption{Normalized score across all 9 \IsaacGym tasks that we consider, aggregated using the IQM. \RLE achieves a higher interquartile mean of normalized score compared to both variants of \PPO, indicating that it can improve over \PPO in continuous control domains as well. Meanwhile, \RLE performs similarly to \RND in terms of aggregated performance.}
    \label{fig:result:iqm_normalized_score_isaacgym}
\end{figure}

\begin{figure}[htb!]
    \centering
    \includegraphics[width=0.5\textwidth]{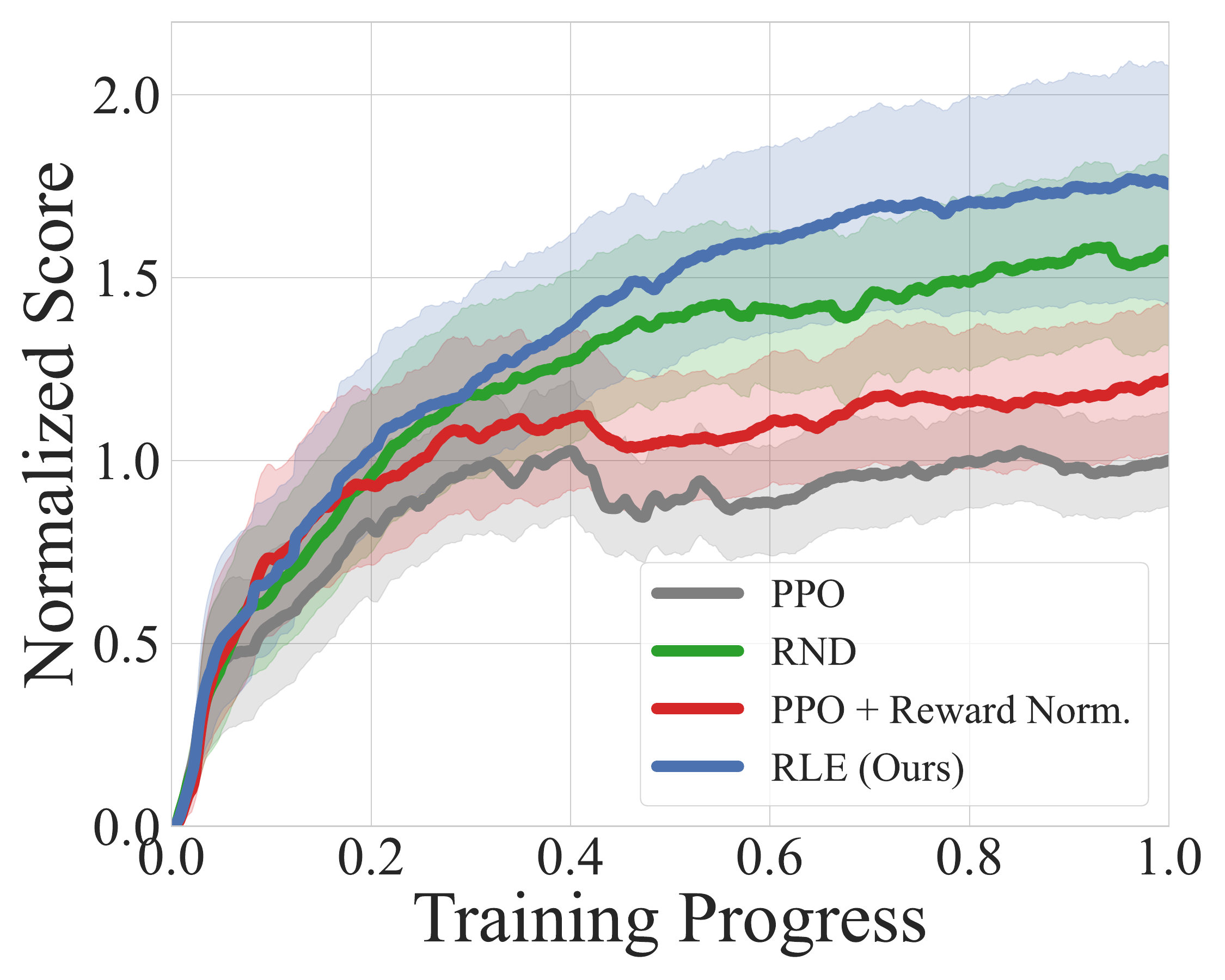}
    \caption{Mean normalized score across all 9 \IsaacGym tasks. Aggregating using the mean yields similar results: \RLE outperforms both variants of \PPO, and slightly outperforms \RND overall.}
    \label{fig:result:mean_normalized_score_isaacgym}
\end{figure}

\begin{figure*}[htb!]
    \centering
    \includegraphics[width=0.6\textwidth]{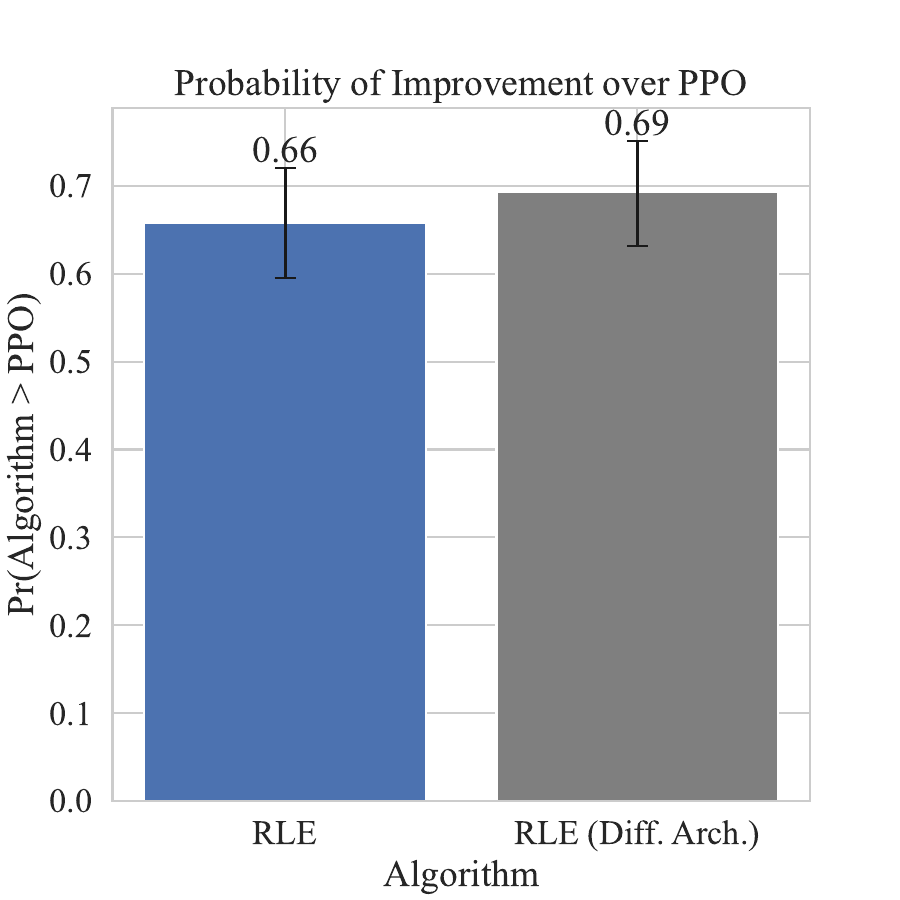}
    \caption{Probability of improvement over \PPO of \RLE and \RLE with a different architecture for $\phi$. The probability of improvement for both \RLE variants is close and the confidence intervals for the probability of improvement metric heavily overlap. This suggests that \RLE is robust to the choice of architecture for $\phi$.} 
    \label{fig:result:poi_diff_arch}
\end{figure*}

\begin{figure*}[htb!]
    \centering
    \includegraphics[width=0.5\textwidth]{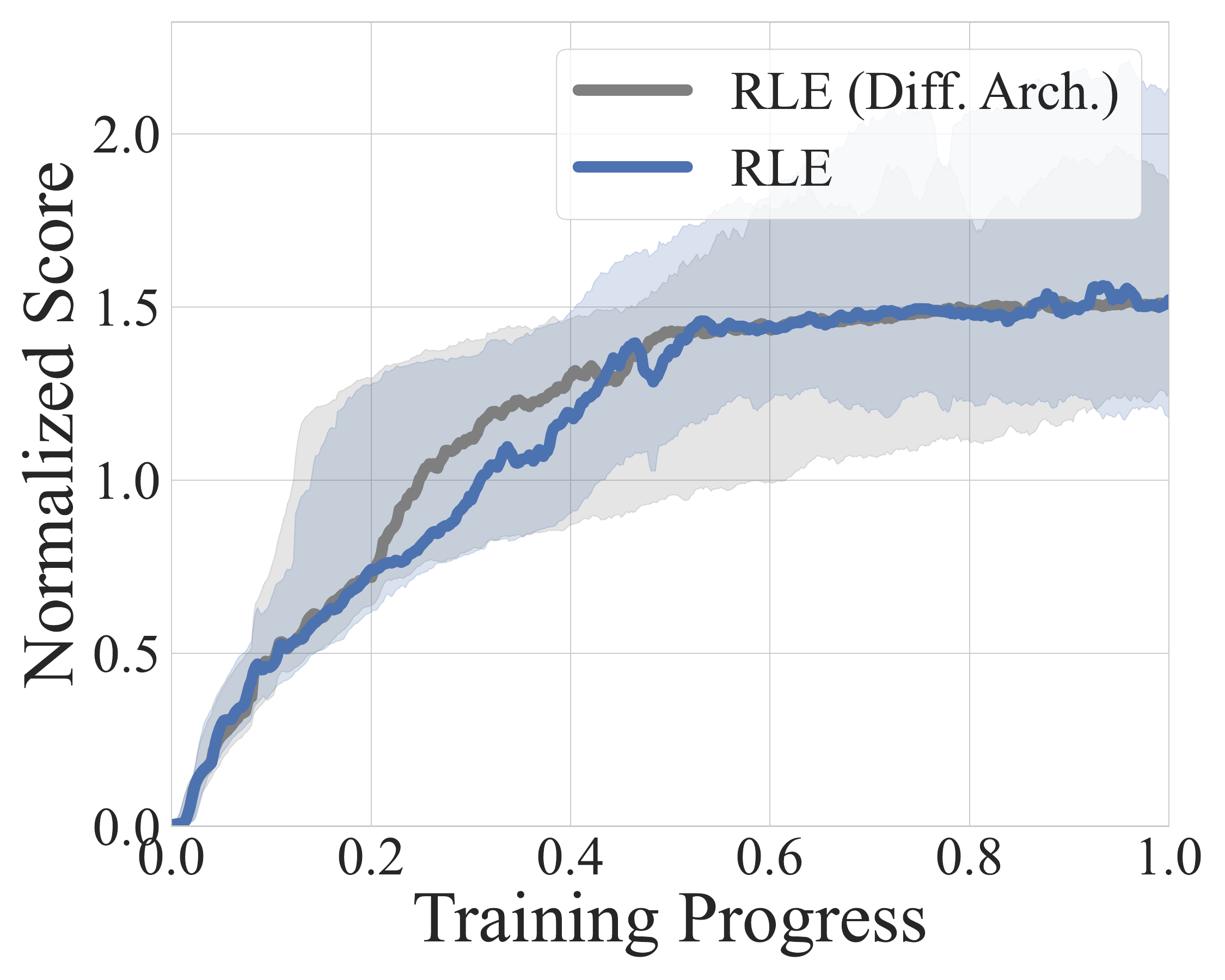}
    \caption{IQM of \PPO normalized score of \RLE and \RLE with a different architecture for $\phi$. The different architecture used in this experiment has less width and uses one less layer. The IQM of normalized score is similar for both methods, suggesting that \RLE does not highly depend on the architecture of the network $\phi$.} 
    \label{fig:result:iqm_diff_arch}
\end{figure*}

\begin{figure*}[htb!]
    \centering
    \includegraphics[width=0.85\textwidth]{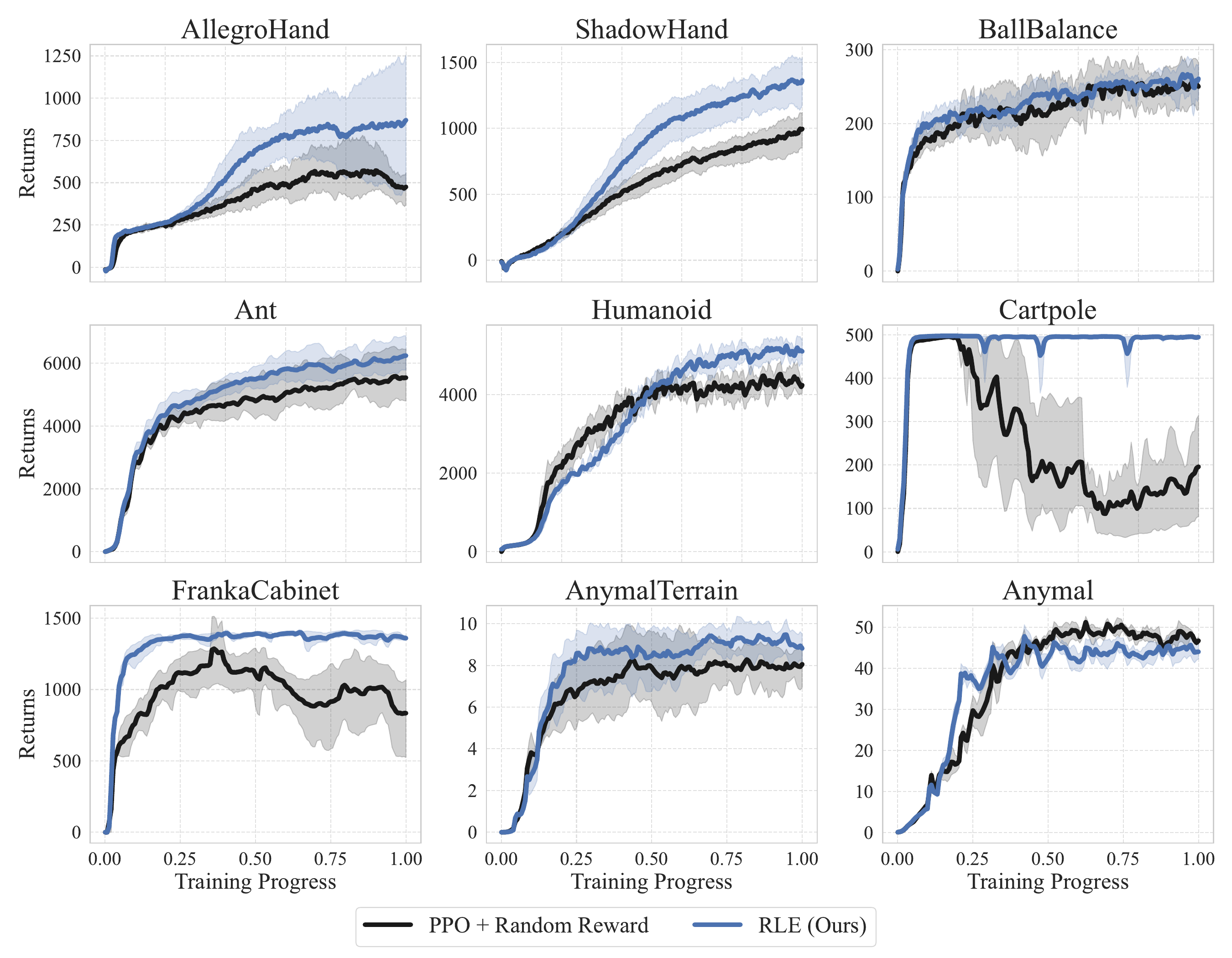}
    \caption{Comparison of achieved returns between \RLE and \PPO with random normal noise sampled i.i.d from a standard normal distribution added to the reward at each timestep. The intrinsic reward coefficient is $0.01$. \RLE outperforms this variant of \PPO in a large majority of games, suggesting that \RLE benefits from using state-dependent random rewards.} 
    \label{fig:result:isaacgym_noise_ablation}
\end{figure*}

We provide: 
\begin{itemize}
    \item The learning curves for \PPO and \RLE in all 9 \IsaacGym tasks that we consider in \Cref{fig:result:all_isaacgym}. Note that in \textsc{Cartpole}, \PPO performance degrades abruptly in the middle of training, while \RLE maintains high performance throughout, suggesting that \RLE prevents the learning process from collapsing during training.
    \item The POI of all methods over \PPO over all 9 \IsaacGym tasks in \Cref{fig:result:poi_over_ppo}.
    \item The IQM of \PPO normalized score of \RLE, \PPO, and \RND aggregated across all 9 \IsaacGym tasks in \Cref{fig:result:iqm_normalized_score_isaacgym}.
    \item The mean of \PPO normalized score of \RLE, \PPO, and \RND aggregated across all 9 \IsaacGym tasks in \Cref{fig:result:mean_normalized_score_isaacgym}.
    \item An ablation study of how different network architectures for $\phi$ affect performance on \IsaacGym tasks  IQM of \PPO normalized score, and probability of improvement over \PPO). We plot the IQM of \PPO normalized score in \Cref{fig:result:iqm_diff_arch}, and probability of improvement over \PPO in \Cref{fig:result:poi_diff_arch}.
    \item An ablation study of how using white noise for randomizing rewards affects performance, shown in \Cref{fig:result:isaacgym_noise_ablation}.
\end{itemize}

